\documentclass[lettersize,journal]{IEEEtran}
\usepackage{amsmath,amsfonts}
\usepackage{algorithmic}
\usepackage{algorithm}
\usepackage{array}
\usepackage[caption=false,font=normalsize,labelfont=sf,textfont=sf]{subfig}
\usepackage{textcomp}
\usepackage{stfloats}
\usepackage{url}
\usepackage{verbatim}
\usepackage{pifont}
\usepackage{marvosym}
\usepackage{graphicx}
\usepackage{enumitem}
\usepackage{cite}
\usepackage{amsfonts}
\usepackage{booktabs}
\usepackage{multirow}
\usepackage{makecell}
\usepackage{color}
\usepackage{forest}
\usetikzlibrary{arrows.meta,shapes,positioning,shadows,trees}
\usepackage{lscape}
\usepackage{orcidlink}

\usepackage{xcolor}

\definecolor{deep-pink}{rgb}{1.0, 0.08, 0.58}

\usepackage{hyperref}
\usepackage{marvosym}
\usepackage{graphicx}
\usepackage{listings}
\definecolor{mycolor}{HTML}{FF6666}

\begin{document}

\twocolumn

\title{Bridging Evolutionary Algorithms and Reinforcement Learning: A Comprehensive Survey on Hybrid Algorithms}
% \,\orcidlink{0000-0002-4667-3505}

\author{
Pengyi Li\,\orcidlink{0009-0009-8546-2346}, Jianye Hao\,\orcidlink{0000-0002-0422-8235},~\IEEEmembership{Senior Member,~IEEE}, Hongyao Tang\,\orcidlink{0000-0001-7478-7684}, Xian Fu\,\orcidlink{0000-0002-4667-3505}, Yan Zheng\,\orcidlink{0000-0002-5016-6549}, \\
Ke Tang\,\orcidlink{0000-0002-6236-2002},~\IEEEmembership{Fellow,~IEEE}
        % <-this % stops a space
\thanks{This work is supported by the National Natural Science Foundation of China (Grant Nos. 92370132, 62106172), the Xiaomi Young Talents Program of Xiaomi Foundation, the National Project X of China (Grant No. JCKY2021204B104), and the Science and Technology on Information Systems Engineering Laboratory (Grant Nos. WDZC20235250409, 6142101220304). (Corresponding
author: Jianye Hao.)}

%\thanks{This work is supported by the National Natural Science Foundation of China (Grant Nos. 92370132). (Corresponding author: Jianye Hao.)}

%Corresponding authors: Jianye Hao ( \href{mailto:jianye.hao@tju.edu.cn}{jianye.hao@tju.edu.cn})}
\thanks{
Pengyi Li, Jianye Hao, Xian Fu, and Yan Zheng are with the College of Intelligence and Computing, Tianjin University, Tianjin 300350, China (e-mail: \href{mailto:lipengyi@tju.edu.cn}{lipengyi@tju.edu.cn};
 \href{mailto:jianye.hao@tju.edu.cn}{jianye.hao@tju.edu.cn};
 \href{mailto:xianfu@tju.edu.cn}{xianfu@tju.edu.cn};
 \href{mailto:yanzheng@tju.edu.cn}{yanzheng@tju.edu.cn}
 ).

Hongyao Tang is with the Montreal Institute of Learning Algorithms (MILA), Quebec H3A 0G4, Canada (e-mail: \href{mailto:tang.hongyao@mlia.quebec}{tang.hongyao@mlia.quebec}

Ke Tang is with the Department of Computer Science and Engineering, Southern University of Science and Technology, Shenzhen 518055, China (e-mail: \href{mailto: tangk3@sustech.edu.cn}{tangk3@sustech.edu.cn})
 }% <-this % stops a space

}

\maketitle

\begin{abstract}
Evolutionary Reinforcement Learning (ERL), which integrates Evolutionary Algorithms (EAs) and Reinforcement Learning (RL) for optimization, has demonstrated remarkable performance advancements. By fusing both approaches, ERL has emerged as a promising research direction. This survey offers a comprehensive overview of the diverse research branches in ERL.
Specifically, we systematically summarize recent advancements in related algorithms and identify three primary research directions: EA-assisted Optimization of RL, RL-assisted Optimization of EA, and synergistic optimization of EA and RL. Following that, we conduct an in-depth analysis of each research direction, organizing multiple research branches. We elucidate the problems that each branch aims to tackle and how the integration of EAs and RL addresses these challenges. In conclusion, we discuss potential challenges and prospective future research directions across various research directions.
To facilitate researchers in delving into ERL, we organize the algorithms and codes involved on \textcolor{deep-pink}{\href{https://github.com/yeshenpy/Awesome-Evolutionary-Reinforcement-Learning}{https://github.com/yeshenpy/Awesome-Evolutionary-Reinforcement-Learning}}
% 我们讨论了不同研究方向存在的潜在挑战以及潜在的未来研究方向
\end{abstract}

\begin{IEEEkeywords}
Evolutionary Algorithms, Reinforcement Learning, Evolutionary Reinforcement Learning.
\end{IEEEkeywords}

\section{Introduction}
\label{Intro}

% 强化学习近些年来收到了大量的关注并取得了重大的发展。
%Reinforcement learning (RL)~\cite{sutton2018reinforcement} has received considerable attention and made significant advancements in recent years.
% 深度学习的发展给强化学习带来了新的生命力。

%强化学习是机器学习领域的一类重要的学习方法，擅长于解决各种序列决策问题。RL通常需要将优化问题建模为MDP，并基于贝尔曼方程通过各类优化方式（e.g., 梯度下降）进行策略搜索。Thanks to深度学习的发展，强化学习的能力得到了进一步提升。通过利用神经网络强大的表达能力以及梯度优化高效的优化效率，RL能够拟合更加复杂的价值函数，同时展示出了更高地优化效率相较于无梯度优化方法。
\IEEEPARstart{R}{einforcement} Learning (RL)~\cite{sutton2018reinforcement} is an important category of learning methods within the field of machine learning~\cite{jordan2015machine}, specializing in solving various sequential decision-making problems.
By modeling sequential decision-making problems as Markov Decision Processes (MDPs)~\cite{puterman1990markov}, RL learns an optimal policy via iterative policy optimization and evaluation with various optimization techniques (e.g., gradient descent~\cite{andrychowicz2016learning}).
%RL needs to model optimization problems as Markov Decision Processes (MDPs)~\cite{puterman1990markov} and perform policy optimization based on the Bellman equation~\cite{sutton2018reinforcement} through various optimization techniques (e.g., gradient descent~\cite{andrychowicz2016learning}). 
Thanks to the development of deep learning, the capabilities of RL have been further enhanced. 
By leveraging the powerful expressive capabilities of neural networks and efficient gradient optimization, RL can approximate more complex value functions and demonstrate superior learning efficiency compared to gradient-free methods~\cite{GangwaniP18GPO}.
In addition, RL, especially off-policy RL, collects and reuses historical samples, significantly improving sample efficiency and making it more applicable to problems where samples are costly~\cite{redq}. 
With these advancements, RL has achieved significant successes in diverse domains, including Game AI~\cite{vinyals2019grandmaster}, Robotics~\cite{johannink2019residual}, Recommender System~\cite{zou2019reinforcement}, and Scheduling Problems~\cite{NiHLTYDM021MultiGraph}.
However, RL still faces several inherent and long-standing challenges, including limited exploration abilities~\cite{hao2023exploration}, poor convergence~\cite{survey2}, sensitivity to hyperparameters~\cite{Hyperparameters}, and suboptimality in gradient optimization~\cite{Primacy}. 
These challenges hinder the application of RL in more complex real-world scenarios.
% 这些问题潜在地限制了RL的性能边界，从而阻碍了RL应用到更加复杂的现实场景中。
%These challenges prevent RL from being sample efficient enough when applied to real-world problems and restrict its practical applicability in real-world problems.
%Additionally, RL typically necessitates the modeling of the problem as a Markov Decision Process (MDP), which often introduces additional problems and challenges when applying RL to other types of optimization problems~\cite{survey1}.
% 这使得RL在解决其他优化问题时往往会引入一些额外的问题与挑战。
% thereby making its application to single-state optimization problems challenging~\cite{survey1}. 

Evolutionary Algorithms (EAs)~\cite{back1993overview, ea-book, vikhar2016evolutionary, zhou2019evolutionary, DBLP:conf/gecco/Jong20} are a class of gradient-free, black-box optimization methods. By emulating Darwin's theory of evolution, EAs iteratively evolve solutions. Due to the diversity within populations and the gradient-free random search, EAs have strong exploration ability. As a result, compared to conventional local search algorithms like gradient descent, EAs exhibit better global optimization capabilities within the solution space and are adept at solving multimodal problems~\cite{vikhar2016evolutionary, li2010species}. Moreover, EAs show good robustness and convergence properties, displaying resistance to noise and uncertainty~\cite{jin2005evolutionary}. 
With these characteristics, EAs have demonstrated formidable capabilities in various practical optimization problems~\cite{han2002quantum, such2017deep}, including path planning~\cite{yu2020constrained}, scheduling problems~\cite{gao2019review}, and circuit design~\cite{dasgupta2013evolutionary}.
{Besides, EAs have also demonstrated the capacity to evolve a single policy for solving multi-task sequential decision problems~\cite{kelly2018emergent,smith2019model,kelly2021evolving}.}
%Nonetheless, EAs have notable weaknesses, such as reliance on expert knowledge for operator design and selection~\cite{li2023scheduling}, 
%low learning efficiency~\cite{GangwaniP18GPO}, %inability to be data-driven, and thus unable to leverage existing experiences. 
{Nonetheless, EAs also have weaknesses, including sensitivity to the design and selection of variation operators~\cite{li2023scheduling}, as well as ineffective and redundant exploration arising from random search~\cite{GangwaniP18GPO}. Besides, EAs often demonstrate low sample efficiency in sequential decision-making problems~\cite{survey2, wang2022surrogate, Re2}.}

\tikzset{
    root/.style = {draw=pink, text width=1cm,  align=center, rectangle, rounded corners},
    anode/.style = {draw=pink, text width=3cm,  align=center, rectangle, rounded corners,},
    bnode/.style = {draw=pink, text width=3cm, align=center, rectangle, rounded corners,},
    cnode/.style = {draw=pink, text width=12cm,  align=left, rectangle, rounded corners, },
    edge from parent/.style={draw=black, edge from parent fork right},
}

{\begin{figure*}
\centering
\resizebox{0.9\textwidth}{!}{
\begin{forest}
for tree={
    %font=\normalfont\ttfamily,
    grow=east,
    growth parent anchor=east,
    parent anchor=east,
    child anchor=west,
    edge path={\noexpand\path[\forestoption{edge},-, >={latex}] 
         (!u.parent anchor) -- +(5pt,0pt) |- (.child anchor)
         \forestoption{edge label};},
    %  s sep=0cm, % 设置同一层级的兄弟节点之间的水平距离
    % calign=child center, % 将父节点指向子节点文本框的中心锚点
    anchor=center, % 设置文本框的中心锚点为节点的锚点
    draw=pink,
    if n children=0{
        fill=pink,
        text=black,
        %edge={pink!30},
    }{
    }
}
[ERL, root%,  rotate=90
    [Synergistic Optimization of EA and RL, anode   
        [Learning Classifier Systems , bnode
        [{XCS~\cite{XCS}, XCSG~\cite{XCS},XCSF~\cite{XCSF}, XCSF w/ tile coding~\cite{XCSFtile}, DGP-XCSF~\cite{DPGXCSF}}, cnode, ]
        ]
         [Interpretable AI, bnode
            [{POC-NLDT~\cite{dhebar2020interpretable}, GE-QL~\cite{custode2022interpretable}, CG-DT~\cite{custode2022interpretable}, CC-POC~\cite{custode2021co}, QD-GT~\cite{ferigo2023quality}, SIRL~\cite{lucio2024social}, SVI~\cite{kubalik2021symbolic}}, cnode, ]
        ]
        [Morphological Evolution, bnode
            [{GA (EvoGym)~\cite{EvoGym}, HERD~\cite{HERD}, AIEA~\cite{liu2023rapidly}, DERL~\cite{Fifei}, TAME~\cite{hejna2021task}, Others~\cite{DBLP:conf/gecco/PigozziVM23}}, cnode, ]
        ]
        [Reward Design, bnode
            [{Evo-Reward~\cite{Evo-Reward}, Eureka~\cite{Eureka}, DrEureka~\cite{Dreureka}, ROSKA~\cite{ROSKA}, R*~\cite{li2025r}, LaRes~\cite{li2025lares}, ReMAC~\cite{li2025remac}}, cnode, ]
        ]
        [Multi-Agent Optimization, bnode
            [{ MERL~\cite{MERL}, NS-MERL~\cite{NS-MERL}, CEMARL~\cite{CEMARL}, EMARL~\cite{EMARL}, RACE~\cite{li2023race}}, cnode]
        ]
        [Single-Agent Optimization, bnode
            [{ERL~\cite{khadka2018evolution}, CERL~\cite{khadka2019collaborative}, PDERL~\cite{bodnar2020proximal}, SC~\cite{wang2022surrogate}, GEATL~\cite{GEATL}, CSPS~\cite{DBLP:conf/nips/ZhengW0L0Z20}, T-ERL~\cite{DBLP:conf/gecco/ZhengC23}, ESAC~\cite{DBLP:conf/atal/Suri22}, PGPS~\cite{kim2020pgps}, ERL-Re$^2$~\cite{Re2}, VEB-RL~\cite{li2023value}, EvoRainbow-Exp~\cite{lievorainbow} , EvoRainbow~\cite{lievorainbow}, CORE~\cite{li2025core}}, cnode, ]
        ]
    ]
    [RL-assisted Optimization of EA, anode
        [Others, bnode
        [{Grad-CEM~\cite{Grad-CEM}, LOOP~\cite{LOOP}, TD-MPC~\cite{TD-MPC}, RGP~\cite{RGP}, GNP-RL~\cite{GNP-RL}, LPO~\cite{lu2022discovered},  TA-LPG~\cite{jackson2024discovering}, TA-LPO~\cite{jackson2024discovering}}, cnode, ]
        ]
        [Hyperparameter Configuration, bnode
            [{AGA~\cite{DBLP:conf/esoa/EibenHKS06}, LTO~\cite{LTO}, RL-DAC~\cite{DBLP:conf/ecai/BiedenkappBEHL20}, REM~\cite{zhang2022variational}, Q-LSHADE \&DQ-HSES~\cite{DBLP:journals/cim/ZhangSBZX23}, MADAC~\cite{DBLP:conf/nips/0001XYL0Z022}, qlDE~\cite{qlDE}, RLDE~\cite{hu2021reinforcement}}, cnode,]
        ]
        [Dynamic Operator Selection, bnode
            [{RL-GA(a)~\cite{DBLP:conf/gecco/PettingerE02}, RLEP~\cite{RLEP}, EA+RL~\cite{EA+RL}, EA+RL(O)~\cite{EA+RL(O)}, RL-GA(b)~\cite{DBLP:journals/swevo/SongWYWXC23}, GSF~\cite{yi2022automated}, MARLwCMA~\cite{DBLP:journals/access/SallamECR20}, MPSORL~\cite{meng2023multi}, DEDQN ~\cite{DEDQN}, DE-DDQN~\cite{sharma2019deep}, RL-CORCO~\cite{RL-CORCO}, RL-HDE~\cite{RL-HDE}, DE-RLFR~\cite{DBLP:journals/swevo/LiSYSQ19}, LRMODE ~\cite{huang2020afitness}, MOEA/D-DQN~\cite{tian2022deep}, AMODE-DRL~\cite{li2023scheduling}}, cnode]
        ]
        [Variation Operator, bnode
            [{GPO~\cite{GangwaniP18GPO}, CEM-RL~\cite{pourchot2018cem}, CEM-ACER~\cite{tang2021guiding}, PBRL~\cite{pretorius2021population}, NS-RL~\cite{Shi2020ffficient}, DEPRL~\cite{liu2021diversity}, QD-RL~\cite{QDRL}, PGA-ME~\cite{nilsson2021policy}, GAC QD-RL~\cite{GAC-QD-RL}, CMA-MEGA~\cite{tjanaka2022approximating}, CCQD~\cite{anonymous2024sampleefficient}, RefQD~\cite{wang2024quality}, Wuji~\cite{wuji}}, cnode, edge path={
    \noexpand\path [\forestoption{edge}]
    (!u.parent anchor) -- +(0,0pt) -| (.child anchor)\forestoption{edge label};
  },]
        ]
        [Population Evaluation, bnode
            [{SC~\cite{wang2022surrogate}, PGPS~\cite{kim2020pgps}, ERL-Re$^2$~\cite{Re2}}, cnode, ]
        ]
        [Population Initialization, bnode
            [{NGGP~\cite{NGGP}, RL-guided GA~\cite{RL_guided_EA}, DeepACO~\cite{DeepACO}}, cnode,]
        ]
    ]
    [EA-assisted Optimization of RL, anode
        [Others, bnode
            [{Evo-Reward~\cite{Evo-Reward}, DQNClipped \& DQNReg~\cite{EvoRL}, GP-MAXQ~\cite{GP-MAXQ}, PNS-RL~\cite{liu2021population}, Go-Explore~\cite{Go-exp}, G2N~\cite{chang2018genetic}, EVO-RL~\cite{hallawa2021evo}, ROMANCE~\cite{ROMANCE}, MA3C~\cite{MA3C}, EPC~\cite{EPC}, MAPPER~\cite{MAPPER}}, cnode, ]
        ]
        [Hyperparameter Optimization, bnode
            [{OMPAC~\cite{elfwing2018online}, PBT~\cite{PBT}, SEARL~\cite{franke2021sample}, GA-DRL~\cite{sehgal2022ga}, AAC~\cite{grigsby2021towards}, OHT-ES~\cite{tang2020online}}, cnode, ]
        ]
        [EA-assisted Action Selection, bnode
            [{Qt-Opt~\cite{kalashnikov2018scalable}, CGP~\cite{simmons2019q}, EAS-RL~\cite{ma2022evolutionary}, SAC-CEPO~\cite{shi2021soft}, GRAC~\cite{shao2022grac}, OMAR~\cite{OMAR}, COMIX~\cite{COMIX}}, cnode, ]
        ]
        [EA-assisted Parameter Search, bnode
            [{EQ~\cite{leite2020reinforcement}, Supe-RL~\cite{marchesini2021genetic}, VFS~\cite{marchesini2023improving}}, cnode, ]
        ]
    ]       
]
\end{forest}
}
\caption{Three major research directions in the ERL field. Each direction comprises multiple research branches.}
\label{survey list}
\end{figure*}
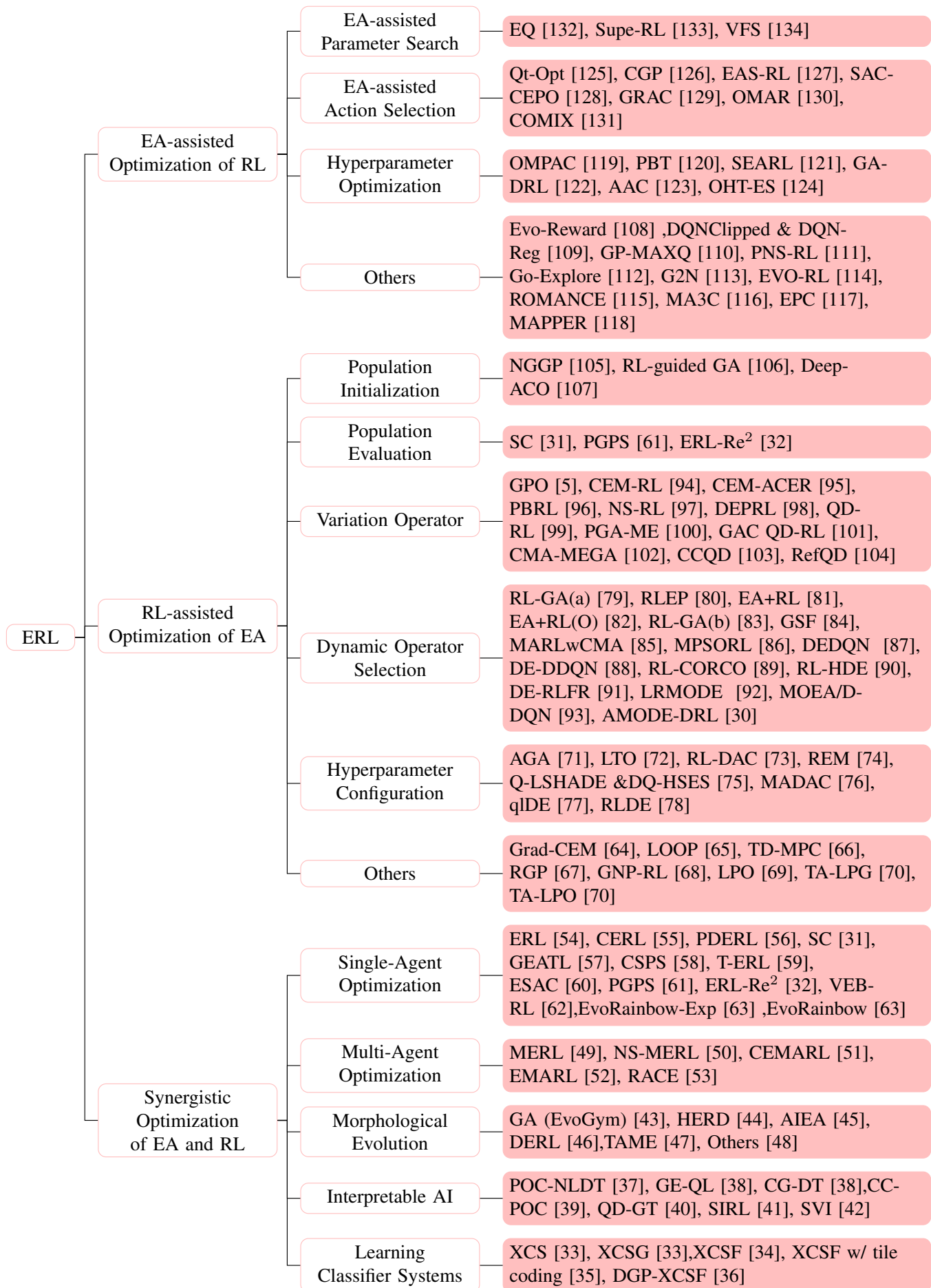

As discussed above, RL and EAs are two categories of methods based on different principles. {Each has proven its efficiency in addressing specific problems but also faces different challenges.
% Addressing these challenges is crucial for further enhancing the efficiency in solving various practical problems. 
With the development of the EAs and RL communities, many researchers have found that these challenges can be addressed by combining advanced methods from both fields, leading to the emergence of numerous hybrid methods. For brevity, we refer to these works as Evolutionary Reinforcement Learning (ERL).
However, the technical pathways of these methods and the problems they aim to address are various and diverse. There is a lack of a comprehensive survey and systematic analysis in the ERL literature. 
In this paper, we attempt to provide a systematic review of existing ERL works. We first revisit these works from the following three primary perspectives:} %RL, EAs, and collaboration of EAs and RL.
%as these perspectives are more closely aligned with the technical pathways involved. 

{\large{\textcircled{\small{1}}}\normalsize \ \textbf{From the perspective of RL}: RL is a class of learning algorithms used to tackle various sequential decision-making problems. Apart from modeling problems as MDPs, RL involves configuring algorithms, interaction, and learning and optimizing network parameters. This process encounters many optimization challenges, such as hyperparameter optimization, action selection, and network parameter optimization. EAs exhibit strong search capabilities and global optimization prowess, which can further enhance the quality of solving such optimization sub-problems within RL.
% For instance, utilizing EAs to assist in hyperparameter optimization for RL.
}

{\large{\textcircled{\small{2}}}\normalsize \  \textbf{From the perspective of EAs}:
EAs are a category of optimization algorithms that require iterative searches in the solution space to obtain feasible solutions. This typically involves population initialization, evaluation, operator design and selection, and algorithm configuration. However, this process often confronts assorted hurdles, such as how to construct the initial population, which often determines the quality of the final solution; how to perform efficient mutation that avoids redundant and inefficient exploration; and how to dynamically select operators at different stages of optimization to improve performance. The learning ability of RL enables it to develop strong discriminative guidance and decision-making capabilities. Incorporating RL into the optimization process of EAs has been proven to further enhance the efficiency of EAs.}

{\large{\textcircled{\small{3}}}\normalsize \ \textbf{From the perspective of collaboration}: EAs and RL can collaborate to solve a problem, typically through two approaches: 1) EAs and RL simultaneously address the same problem and collaborate through some mechanisms. 2) EAs and RL each solve a part of the problem and eventually integrate to form a complete solution. The former approach aims to complement the strengths and weaknesses of EAs and RL in problem-solving: EAs' exploration capability compensates for RL's exploration limitations, while RL's experience reuse and fine-grained learning address EAs' sample inefficiency. The latter approach involves EAs and RL tackling tasks they excel in individually; for instance, EAs optimizing topology and RL learning control policies.}

{Based on the insights mentioned above,}
we categorize the ERL works into three main directions:
1) \textbf{EA-assisted Optimization of RL}. 
This integration approach incorporates EAs into the learning process of RL, mainly applied to address sequential decision-making problems.
It leverages diverse exploration, global optimization capabilities, and strong convergence and robustness of EAs to address challenges in RL such as limited exploration capabilities, sub-optimality in gradient optimization, and sensitivity to hyperparameters.
2) \textbf{RL-assisted Optimization of EA}. Opposite to the previous category, this integration approach incorporates RL into the optimization flow of EAs,  mainly applied to address various optimization problems and sequential decision-making problems. It leverages efficient experience utilization and learning capabilities, along with the discriminative abilities of RL to address challenges in EAs such as population initialization, algorithm configuration, uncontrollable mutation, and high sample cost in fitness evaluation. 
3) \textbf{Synergistic Optimization of EA and RL}.
This integrated approach maintains complete processes for both EAs and RL to collaboratively address the same problem simultaneously or independently tackle partial solutions, which are subsequently integrated into a complete solution.
This allows each of them to leverage their respective strengths in problem-solving and mutually enhance each other through complementary features, ultimately achieving better performance.
%The related works in this direction primarily focus on sequential decision-making problems.
% \textcolor{red}{For instance, EAs and RL search solutions using policy gradient guidance and population evolution, respectively. Additionally, EAs leverage population-based exploration to address the limited exploration capability of RL. RL, on the other hand, can reuse samples generated by EAs to enhance sample efficiency.} \thy{need to polish the expression}
%For instance, EAs and RL search policies simultaneously. During this process, EAs share experiences generated during the evaluation process with RL to overcome the limited exploration capability. Conversely, RL injects the optimized policy back into the population to improve the evolutionary efficiency of EAs.
% EAs与RL分别通过策略梯度优化以及演化来进行解的学习，在这个学习过程中，EAs将学习过程中产生的经验提供给RL来解决RL探索能力不足的问题，RL则将优化后的个体注入到种群中来加速演化。
%

{There have been some efforts to review works related to EAs and RL, including comparing EAs with RL algorithms~\cite{ma_ea_survey}, exploring the integration of EAs and RL for policy search~\cite{survey2, ZHU2023126628}, reviewing the hybrid algorithms within a unified framework, including motivation, natural models, sub-algorithms, techniques, and properties~\cite{survey1}, 
reviewing hybrid algorithms based on the challenges they address in RL~\cite{kelly2023evolutionary},
and reviewing hybrid algorithms according to the key research areas of RL~\cite{survey3}.}
However, these surveys lack comprehensive and systematic investigation into hybrid algorithms. Thus we aim to furnish a more systematic and comprehensive survey to fill this gap. We summarize our contributions as follows:
\begin{itemize}
    \item We conduct a thorough and systematic analysis of the works in the ERL domain, leading to the establishment of three research directions: EA-assisted Optimization of RL, RL-assisted Optimization of EA, and Synergistic Optimization of EA and RL.
    \item In each direction, we further subdivide the works into distinct research branches. Subsequently, we conduct an in-depth analysis of the fundamental issues to be addressed and the related algorithms for each branch.
    \item Furthermore, we point out open challenges within each domain and propose potential research directions to address these challenges.
\end{itemize}

We begin by providing an overview of the basic algorithms in EAs and RL and definitions of involved problems in Section~\ref{sec: background}. 
Subsequently, we delve into each direction to categorize the related works and present them from various branches, outlining the specific problems they address and the proposed approaches. After presenting the various branches within each research direction, we summarize open challenges and discuss potential future directions.

\section{Background}
\label{sec: background}
\subsection{Evolutionary Algorithms}
%Evolutionary algorithms: A critical review and its future prospects

\begin{figure}[t]
\centering
\includegraphics[width=1.0\linewidth]{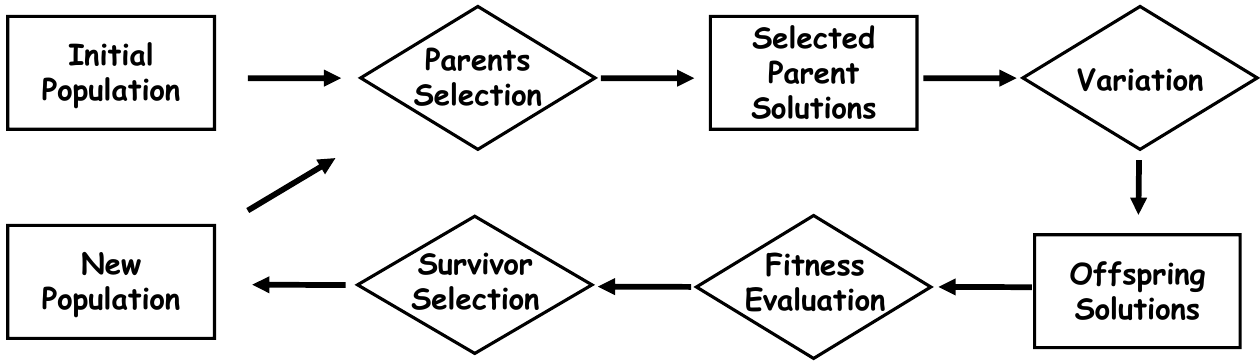}
%\vspace{-0.2cm}
\caption{Evolutionary Algorithm optimization process. Diamond-shaped blocks represent the actions taken by the algorithm, while rectangular blocks represent instances generated by the actions.}
\label{Sec2: EA optimization process}
%\vspace{-0.2cm}
\end{figure}
% 菱形方块代表算法采取的动作，矩形方块代表动作产生的实例

Evolutionary Algorithms (EAs) are a class of biologically inspired gradient-free optimization methods that emulate biological evolution processes~\cite{back1993overview, vikhar2016evolutionary, DBLP:conf/gecco/Jong20}. Figure \ref{Sec2: EA optimization process} illustrates the entire evolutionary process. EAs begin by initializing the population with an initial set of individuals $\mathbb{P}=\{I_1, I_2, ..., I_{N}\}$. These individuals can have various forms in different problems, such as vectors, neural networks, and so on. Subsequently, EAs select parents using a selection operator, and offspring individuals are generated through variation. The selection operator typically involves choosing individuals with the highest fitness scores determined through evaluation. Following this, the offspring are evaluated for their fitness, followed by survivor selection. The selected individuals form the new population for the next generation. 
All the EAs can be abstracted into the aforementioned process.
%EAs have gained recognition as effective methods for a wide range of optimization problems. 
In this survey, we cover various EAs~\cite{DBLP:conf/gecco/Jong20}, such as Genetic Algorithm (GA)~\cite{mitchell1998introduction, such2017deep}, Differential Evolution (DE)~\cite{DBLP:journals/tec/DasS11}, Evolution Strategy (ES)~\cite{salimans2017evolution}, Novelty Search (NS)~\cite{lehman2011novelty}, MAP-Elites~\cite{mouret2015illuminating, nilsson2021policy}, Genetic Programming (GP)~\cite{koza1994genetic}, among others. 
{Fixed genomes are a common feature across GA, DE, and ES. In ES and DE, genomes are typically represented as real-valued vectors, whereas GA utilizes binary strings. Recently some works have expanded the forms of these methods' genomes, allowing them to employ neural networks as genomes~\cite{survey2}. While NS and MAP-Elites are two diversity mechanisms. GP stands out for its ability to handle variable-length genomes, such as tree structures, which enables the exploration of complex topology.}
{It is worth noting that in sequential decision-making problems, existing EAs typically utilize the policies, i.e., neural networks, as individuals in the population, and rely on these policies for action decision-making and interaction. Ultimately, they evaluate the population based on the averaged cumulative rewards obtained throughout several episodes of games, followed by evolution in the parameter space~\cite{pourchot2018cem, Re2}.
We can observe that EAs struggle to utilize finer-grained information, such as states, actions, and step-level rewards. This is one major factor leading to sample inefficiency.
}

\begin{figure}[t]
\centering
\includegraphics[width=1.0\linewidth]{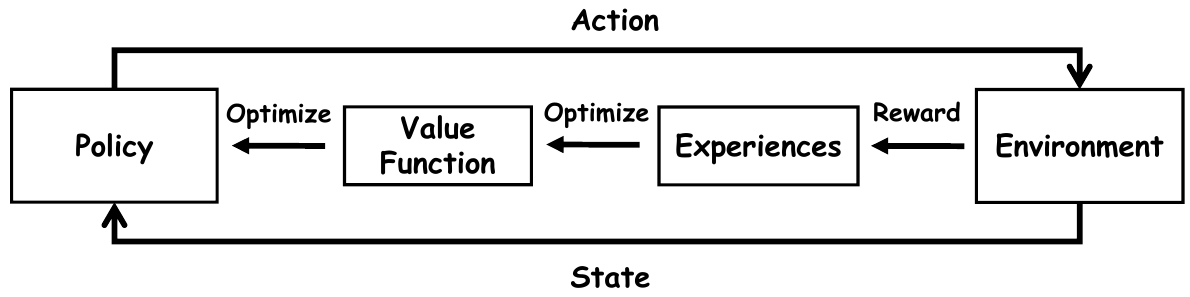}
%\vspace{-0.2cm}
\caption{Reinforcement Learning process. %This figure primarily illustrates the optimization process of off-policy RL algorithms. In contrast, on-policy RL algorithms can omit certain steps, such as the replay buffer.
}
\label{Sec2: RL optimization process}
%\vspace{-0.2cm}
\vspace{-0.3cm}
\end{figure}

\subsection{Reinforcement Learning}
Reinforcement Learning (RL) needs to formalize the target problem as Markov Decision Processes (MDPs)~\cite{sutton2018reinforcement}, where the agent interacts with the environment over several finite time steps. The MDP can be defined by a tuple $(\mathcal{S}, \mathcal{A}, \mathcal{P}, \mathcal{R}, \gamma, T)$, where $\mathcal{S}$ denotes the state space, $\mathcal{A}$ is the action space, $\mathcal{P}: \mathcal{S} \times \mathcal{A} \times \mathcal{S} \rightarrow [0, 1]$ is the state transition function, $\mathcal{R}: \mathcal{S} \times \mathcal{A} \rightarrow \mathbb{R}$ is the reward function, $\gamma$ is the discount factor, and $T$ is the maximum episode length. At each time step $t$, the agent chooses an action $a_t \in \mathcal{A}$ according to the state $s_t \in \mathcal{S}$ and its policy $\pi(s): \mathcal{S} \rightarrow{} \mathcal{A}$, receives a reward $r_t \in \mathcal{R}(s_t, a_t)$, and gets the next state $s_{t+1}$ based on $\mathcal{P}(s_{t+1}|s_t, a_t)$. RL aims to find a policy that maximizes the cumulative discounted rewards $R_t = \sum_{i=t}^{T} \gamma^{i-t}r_i$ at each time step $t$.
%强化学习可以分为value based的RL算法和policy based的RL算法。
%其中value based RL算法中最经典的算法是 Q-learning，The core idea of Q-learning is to use a Q-function $Q(s,a)$ to learn the Q-value, i.e., the cumulative reward obtained from taking a specific action $a$ in a given state $s$. Q-learning selects actions based on the maximum Q-value $\pi = \arg\max_a Q(s,a)$ and updates the Q-function at each step $t$ based on the reward signals $r_t$. 
% 而policy based的RL算法directly optimize the policy of an agent.
%最经典是REINFORCE，
Q-learning~\cite{watkins1992q} is a classical value-based RL algorithm that uses a Q-function $Q(s,a)$ to learn the Q-value, i.e., the cumulative reward obtained from taking a specific action $a$ in a given state $s$. Q-learning selects actions based on the maximum Q-value $\pi = \arg\max_a Q(s,a)$ and employs the Bellman equation to update the Q-function towards the target $y$ using
$Q(s_t,a_t)\leftarrow Q(s_t,a_t)+\alpha(y-Q(s_t,a_t))$
and $y=\sum_{t=i}^{i+n-1}\gamma^{t-i} r_t+\gamma^n  \mathop{max}\limits_{a}Q(s_{t+n},a)$, 
% \begin{equation}
% \begin{aligned}
% &Q(s_t,a_t)\leftarrow Q(s_t,a_t)+\alpha(y-Q(s_t,a_t)),&
% \end{aligned}
% \end{equation}
% \begin{equation}
% \begin{aligned}
% &y=\sum_{t=i}^{i+n}\gamma^{t-i} r_t+\gamma  \mathop{max}\limits_{a}Q(s_{t+n},a),&
% \end{aligned}
% \end{equation}
where $n$ denotes the number of steps considered for future rewards, larger values of $n$ improve the accuracy of the value function approximation by incorporating more reward signals. However, this also leads to increased variance. This approach is referred to as $n$-step TD, where $n$ is often set to 1, considering only the one-step reward.
% % 其中较长的n会引入更多的奖励信号，从而提升价值函数拟合准确性，但也会引入更大的方差，这种方式被称为n-step TD，通常n设置为0，即只考虑一步奖励。
Policy Gradient methods~\cite{williams1992simple, kakade2001natural} are a class of policy-based RL algorithms that directly optimize the policy of an agent. The basic idea is to update these parameters in the direction that increases the expected cumulative reward $\Bar{R}_\theta=\mathbb{E}_\tau[R(\tau)]$, which is often done using techniques like gradient ascent. One classical algorithm is REINFORCE~\cite{williams1992simple}, which computes the gradient of $\Bar{R}_\theta$ for the policy parameters using 
$\nabla \Bar{R}_\theta = \frac{1}{M}\sum_{k=1}^M\sum_{t=1}^{T} G_t^k\nabla \log \pi_\theta (a_t^k|s_t^k)$ and $G_t^k=\sum_{i=t}^{T}\gamma^{i-t}r_i^k$,
where $G_t^k$ represents the cumulative discounted rewards from a specific time step $t$ until the end of episode $k$.
However, REINFORCE can only conduct policy improvement after completing one episode, which leads to inefficiency. To solve the problem, the Actor-Critic method (AC)~\cite{konda1999actor} integrates value-based RL into policy gradient updates. This is accomplished by training a critic to supply one-step gradients for the policy, i.e., the actor. %Thanks to the powerful function approximation capability of neural networks, DRL has experienced rapid development in recent years.
The development of deep learning has further enhanced the capabilities of RL, leading to the proposal of many Deep RL algorithms, including DQN\cite{DBLP:journals/corr/MnihKSGAWR13, DBLP:journals/nature/MnihKSRVBGRFOPB15}, DDPG\cite{DBLP:journals/corr/LillicrapHPHETS15}, TD3\cite{DBLP:conf/icml/FujimotoHM18}, SAC\cite{DBLP:conf/icml/HaarnojaZAL18}, TRPO\cite{DBLP:conf/icml/SchulmanLAJM15}, PPO\cite{DBLP:journals/corr/SchulmanWDRK17}, and others.

\begin{figure*}[t]
\begin{center}
\subfloat[]{
\centering
\includegraphics[width=0.24\linewidth]{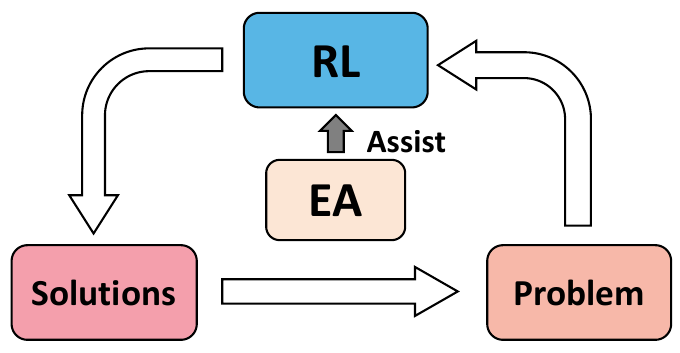}
\label{Figure: EA-Assisted Optimization of RL}}
\subfloat[]{
\centering
\includegraphics[width=0.24\linewidth]{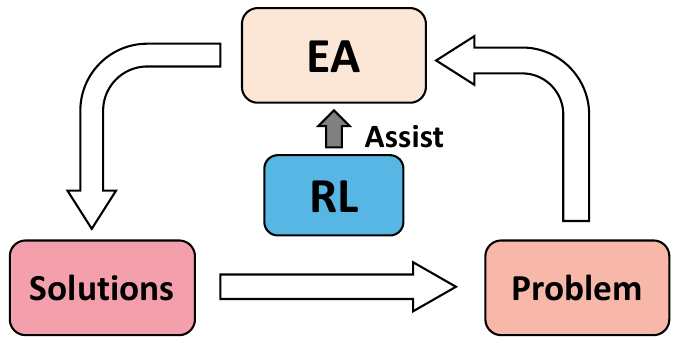}
\label{Figure: RL-Assisted Optimization of EA}}
\subfloat[]{
\centering
\includegraphics[width=0.24\linewidth]{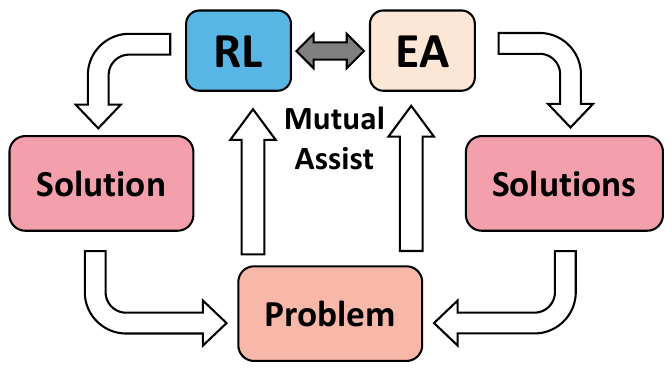}
\label{Schematic for Synergistic Optimization of EA and RL A.}
}
\subfloat[]{
\centering
\includegraphics[width=0.24\linewidth]{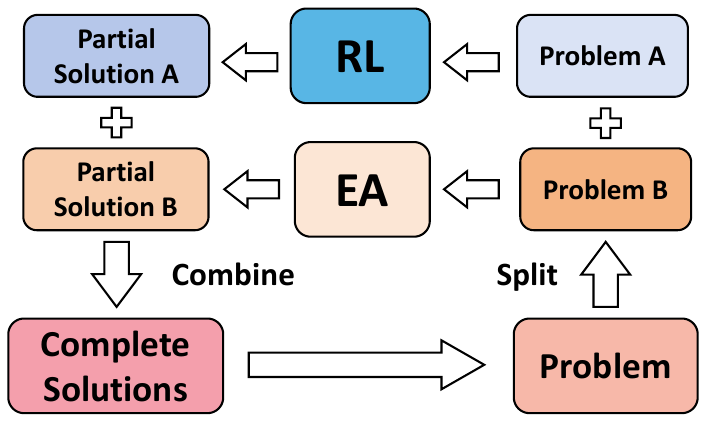}
\label{Schematic for Synergistic Optimization of EA and RL B.}
}
\end{center}
\caption{
Schematic of Four Integration Approaches:
\textit{(a)} EA-assisted Optimization of RL. RL conducts search and improvement of the solution, with EAs playing a supporting role; EAs cannot independently optimize solutions.
\textit{(b)} RL-assisted Optimization of EA. EAs conduct search and improvement of the solution, with RL playing a supporting role; RL cannot independently optimize solutions.
\textit{(c)} and \textit{(d)} Synergistic Optimization of EA and RL.
% RL进行解的搜索与提升，EA为辅助角色，EA不可以单独进行解的优化。
}
\vspace{-0.3cm}
\end{figure*}

\subsection{Problem Definition}
\label{problem defined}
Based on the tasks addressed by the hybrid algorithms, we categorize them into five major classes: sequential decision-making problems, continuous optimization problems, combinatorial optimization problems, multi-objective optimization problems, and multimodal optimization problems. 
%In the context of EA-assisted optimization of RL and Synergistic optimization of EA and RL, the related works primarily focus on sequential decision-making problems. Additionally, in the case of RL-assisted optimization of EA, it also covers the other four categories of optimization problems.
\textbf{Sequential Decision-making Problems (SDP)} involves modeling the task as a MDP. Through policy control, the agent interacts with the environment and receives rewards. The ultimate goal is to obtain a policy that maximizes cumulative rewards. 
Note that other optimization problems can also be modeled as sequential decision problems. For better distinction, we specify the primary tasks involved here: Maze Navigation Problems and Robot Control Problems, including MuJoCo~\cite{todorov2012MuJoCo} and DMC~\cite{DMC}. Additionally, we also explore multi-agent settings, which require training multiple policies to control multiple agents interacting in the environment, aiming to maximize team rewards. The related tasks covered in this paper include SMAC~\cite{QMIX}, MAMuJoCo~\cite{FACMAC}, MPE~\cite{MADDPG}, and flocking tasks~\cite{EMARL}.
In addition, this survey involves the other four types of optimization problems, including \textbf{continuous optimization problems (CTOP)}~\cite{munoz2015algorithm}, \textbf{combinatorial optimization problems (COP)}~\cite{peres2021combinatorial}, \textbf{multi-objective optimization problems (MOOP)}, \textbf{multimodal optimization problems (MMOP)}. CTOP and COP require finding a set of variables $x$ in continuous or discrete spaces to either maximize or minimize the objective function $f(x)$. Typically, such problems come with constraints. CTOP primarily involves the CEC benchmark~\cite{tang2007benchmark, li2013benchmark, kudela2022critical}, while COP involves the Traveling Salesman Problem (TSP)~\cite{flood1956traveling}, Vehicle Routing Problem (VRP)~\cite{toth2002vehicle}, and Scheduling Problems (SP)~\cite{taillard1993benchmarks}, among others. MOOP involves multiple objective functions, rather than a single one. In MOOP, our goal shifts from seeking a solution to minimize or maximize a single objective function, to finding a set of solutions where each represents an optimal solution across different preferences. 
%In single or multi-objective optimization, where a single objective function has multiple local optimal solutions, the objective of MMOP is to find all or as many local sub-optimal solutions as possible.
{Unlike MOOP, MMOP deals with situations in single or multi-objective optimization where a single objective function has multiple local optima. The objective of MMOP is to find all or as many local sub-optimal solutions as possible.}

\begin{table*}[t]
  \renewcommand{\arraystretch}{1.4}
  \centering
  \caption{EA-assisted Optimization of RL.}
  \begin{tabular}{l | lllll}
    \toprule
    \textbf{Class} & \textbf{Algorithm} &  \textbf{Task} & \textbf{EA} & \textbf{RL}  & \textbf{Detail} \\
       \midrule
       
    \multirow{3}{*}{\makecell[l]{Assisted \\Parameter\\ Search}}
    &{EQ~\cite{leite2020reinforcement}} & \makecell[l]{MuJoCo}  & GA & DDPG    & Critic Parameter\\
    \cline{2-6}
    \multirow{1}{*} & {Supe-RL~\cite{marchesini2021genetic}} & \makecell[l]{Navigation \& \\ MuJoCo}& GA & \makecell[l]{PPO \& \\ Rainbow}  & Policy Parameter \\
    \cline{2-6}
    & {VFS~\cite{marchesini2023improving}} & MuJoCo & GA & \makecell[l]{DDPG \& \\PPO \& TD3}  & Critic Parameter \\   
    \hline
    \multirow{6}{*}{\makecell[l]{Assisted \\Action \\ Selection}} & {Qt-Opt~\cite{kalashnikov2018scalable}} & {\makecell[l]{Real Robot Control}}& CEM & Double Q  & Critic Update; Interaction\\
    \cline{2-6}
    \multirow{1}{*} & {CGP~\cite{simmons2019q}} & MuJoCo& CEM & Double Q   & Critic Update; Policy Update \\
    \cline{2-6}
    \multirow{1}{*} & {EAS-RL~\cite{ma2022evolutionary}} &  MuJoCo & PSO & TD3  & Policy Update\\
    \cline{2-6}
    \multirow{1}{*} & {SAC-CEPO~\cite{shi2021soft}} &  MuJoCo & CEM & SAC & Policy Update\\
    \cline{2-6}
    \multirow{1}{*} & {GRAC~\cite{shao2022grac}} & \makecell[l]{ MuJoCo \& \\ Real Robot Control} & CEM & {\makecell[l]{Double \\Q-learning} }  & Critic Update; Policy Update \\
    \cline{2-6}
    \multirow{1}{*} & {OMAR~\cite{OMAR}} & MPE \& D4RL & CEM-like &  \makecell[l]{MA-CQL \\\& CQL}  & Critic Update; Interaction\\
    \cline{2-6}
    \multirow{1}{*} & {COMIX~\cite{COMIX}} & MPE \& MA-MuJoCo & CEM &  \makecell[l]{QMIX}  & Policy Update \\
    % \cline{2-6}
    % \multirow{1}{*} & {PETS} & MuJoCo & CEM &  - & Planning \\
    % \cline{2-6}
    % \multirow{1}{*} & {PlaNet} & MuJoCo & CEM &  - & Planning \\
    % \cline{2-6}
    % \multirow{1}{*} & {POPLIN} & MuJoCo & CEM &  - & Planning \\    
    % \cline{2-6}
    % \multirow{1}{*} & {TD-MPC} & DMC & CEM & DDPG  & Planning \\
    \hline
    % \multirow{1}{*}{PBT} & GA & A3C &  &   &  & \checkmark & \\
    % \midrule
    \multirow{7}{*}{\makecell[l]{Hyperparameter \\ Optimization}}
    \multirow{1}{*} & {OMPAC~\cite{elfwing2018online}} &  Atari \& Tetris & GA & TD($\lambda$) Sarsa($\lambda$)  & $\lambda$, $lr$, $\tau_a$, $\gamma$\\
    \cline{2-6}
    
    \multirow{1}{*} & {PBT~\cite{PBT}} & \makecell[l]{DM Lab \& Atari \& \\ StarCraft II} & GA-like & \makecell[l]{UNREAL \& \\FuN \& A3C}  &  \makecell[l]{$lr$, Entropy Cost, Unroll Length,\\ Intrinsic Reward Cost}\\
    \cline{2-6}
    % & {CERL~\cite{khadka2019collaborative}} & MuJoCo & GA & TD3& $\gamma$\\
    % \cline{2-6}
    \multirow{1}{*} & {SEARL~\cite{franke2021sample}} &  MuJoCo & GA & TD3  & Parameters, $\gamma$, Architecture, Activation \\
    \cline{2-6}

    % \multirow{1}{*}{HOOF} & random search & A2C, TNPG &  &   &  & \checkmark & \\
    % \midrule
    \multirow{1}{*} & {GA-DRL~\cite{sehgal2022ga}} & \makecell[l]{Gym Robotics \& \\ AuboReach} & GA & DDPG & $\gamma, lr, \epsilon$, $\tau,\eta$ (noise std) \\
    \cline{2-6}
    \multirow{1}{*} & {AAC~\cite{grigsby2021towards}} & \makecell[l]{DMC \& \\ Industrial Benchmark}& GA & SAC & \makecell[l]{$\gamma$,  $h$, k, a,c}  \\
    \cline{2-6}
    \multirow{1}{*} & {OHT-ES~\cite{tang2020online}} & DMC & ES & TD3  & $n$, $lr$\\
    % \cline{2-6}
    % \multirow{1}{*} & {DEHB~\cite{awad2021dehb}} & Cartpole Tasks & DE & PPO  & $\gamma, lr,  \epsilon^{clip}, h$, layer size, batch size\\
    \hline
    \multirow{6}{*}{Others} 
    % &  \makecell[l]{Evo-Reward~\cite{Evo-Reward}}& \makecell[l]{Hungry–Thirsty}  & PushGP & Q-learning   &  Reward Function Search\\
    % \cline{2-6}
    &  \makecell[l]{DQNClipped \& \\ DQNReg~\cite{EvoRL}}& \makecell[l]{MiniGrid \& \\  Atari}  & GA-like & DQN   &  Loss Function Search\\
    \cline{2-6}
    \multirow{1}{*}&{GP-MAXQ~\cite{GP-MAXQ}} & \makecell[l]{Foraging Task}  & GP & \makecell[l]{MAXQ}   & 
Hierarchy Search\\
    \cline{2-6}
    & {PNS-RL~\cite{liu2021population}}& \makecell[l]{Gym Robotics \& \\  MuJoCo}  & NS & TD3   & Improve Exploration\\
    \cline{2-6}
    & {Go-Explore~\cite{Go-exp}} & \makecell[l]{Atari}  & GA-like &  Imitation Learning    &  Improve Exploration \\
    \cline{2-6}
    % \multirow{1}{*} & {TD-MPC} & CEM & -   & evolve open-loop controllers \\
    %              \cline{2-6}
    & {G2N~\cite{chang2018genetic}} & \makecell[l]{Atari \& \\  MuJoCo}  & GA & A2C or PPO    &  Activation of neurons \\
    \cline{2-6}
    \multirow{1}{*}&{EVO-RL~\cite{hallawa2021evo}} & Gym Control  & GP & \makecell[l]{Q-learning \& \\ DQN \& PPO}   & Special Setting\\
    \cline{2-6}
    \multirow{1}{*}&{ROMANCE~\cite{ROMANCE}} & SMAC  & QD & \makecell[l]{QMIX}   & Robust MARL\\
    \cline{2-6}
    \multirow{1}{*}&{MA3C~\cite{MA3C}} & \makecell[l]{SMAC, Hallway \\ \& Traffic Junction \\ \& Gold Panner}  & GA-like & \makecell[l]{CMARL}   & Robust Communication\\
    \cline{2-6}
    \multirow{1}{*}&{EPC~\cite{EPC}} & \makecell[l]{Particle-world \\
Environment}  & GA-like & \makecell[l]{MADDPG}   & 
Large-Scale MARL\\
    \cline{2-6}
    \multirow{1}{*}&{MAPPER~\cite{MAPPER}} & \makecell[l]{Grid world}  & GA-like & \makecell[l]{A2C}   & 
Improve Stability\\
    \cline{2-6}
    \multirow{1}{*}&{LPO~\cite{lu2022discovered}} & \makecell[l]{Brax \& MinAtar}  & ES & \makecell[l]{PPO}   & 
Accelerate and Improve Meta RL\\
    \cline{2-6}
    \multirow{1}{*}&{TA~\cite{jackson2024discovering}} & \makecell[l]{Grid world \& Invaders \\ \& Brax \& MinAtar}  & ES & \makecell[l]{LPO \& LPG}   & 
 Aware Learning Time Remaining\\
    % \cline{2-6}
    % & {MPPO} &   MuJoCo & - &  PPO \&  TRPO   &   encourage exploration \\
    % \cline{2-6}
    % & {P3S-TD3} &  MuJoCo  & - &  TD3  & encourage exploration  \\
    \bottomrule
  \end{tabular}
\end{table*}

    % \hline
    % Global &  SEARL~\cite{franke2021sample}, CERLL~\cite{khadka2019collaborative}, PBRL~\cite{pretorius2021population}, FIDI-RL~\cite{shi2019fidi}, CEM-RL~\cite{pourchot2018cem}, CEM-ACER~\cite{tang2021guiding}, PGPS~\cite{kim2020pgps}, DEPRL~\cite{liu2021diversity}, SERL~\cite{wang2022surrogate}, Supe-RL~\cite{marchesini2021genetic}, G2N~\cite{chang2018genetic}  \\
    % \hline
    % Separated & CHDRL~\cite{DBLP:conf/nips/ZhengW0L0Z20}, X-DDPG~\cite{DBLP:conf/mod/EspositiB20a}, Rethinking~\cite{DBLP:conf/gecco/ZhengC23}   \\
    % \hline

% ERL研究方向
% \section{Evolutionary Reinforcement Learning}
% \label{Sec: alg}
% This section provides a comprehensive overview of the three primary directions of ERL: EA-assisted optimization of RL, RL-assisted optimization of EA, and synergistic optimization of EA and RL. We delve into the various branches within each direction, elucidating the specific problems they tackle, and present an integrated introduction to the related algorithms along with their corresponding characteristics. At the end of each subsection, we summarize the challenges and potential directions for future research.

\section{EA-assisted Optimization of RL}

%This subsection offers a comprehensive analysis of the ways in which EA can enhance RL. Based on EA's optimization objective, we categorize the relevant methodologies into three distinct groups: EA-assisted Parameter Search, EA-assisted Action Selection, Hyperparameter Optimization, and Others (Exploration).

% \begin{figure}[t]
% \centering
% \includegraphics[width=1.0\linewidth]{Survey/RL_master.PDF}
% %\vspace{-0.2cm}
% \caption{Schematic for EA-Assisted Optimization of RL.}
% \label{Figure: EA-Assisted Optimization of RL}
% %\vspace{-0.2cm}
% \end{figure}

% 

This section offers a comprehensive analysis of how EAs can enhance RL. 
In this optimization process, the related algorithms revolve around RL, with EAs playing a supporting role in refining this process. EAs cannot solve the problem independently. The optimization schematic is illustrated in Figure \ref{Figure: EA-Assisted Optimization of RL}.
These works primarily focus on sequential decision-making problems, which are the focus of most RL algorithms. %The ultimate goal is to obtain a policy to attain higher cumulative rewards.

According to the optimization objective of EA, we classify the related methods into four branches: EA-assisted Parameter Search, EA-assisted Action Selection, Hyperparameter Optimization, and Others. 
%The works in this chapter primarily focus on Sequential Decision-making Problems, with the ultimate goal of obtaining a policy to accomplish sequential control.
Specifically, 
EA-assisted Parameter Search focuses on leveraging the optimization characteristics of EAs to further improve RL.
EA-assisted Action Selection primarily focuses on the challenge of dealing with a vast action space in continuous action settings, where it is difficult to determine the optimal actions. The goal is to utilize EAs for action search to improve the optimization and decision-making process of RL.
Hyperparameter Optimization focuses on using EAs to automatically adjust RL hyperparameters to mitigate RL hyperparameter sensitivity and improve RL convergence performance.
Others encompass the works that leverage EAs to enhance RL from different perspectives, e.g., reward function search, loss function search, and exploration.
Below, we provide a detailed introduction to each branch and the related works involved.
% 下面我们详细介绍每个分支以及涉及到的工作。

\textbf{EA-assisted Parameter Search.}
% The fundamental goal of Deep Reinforcement Learning (DRL) is to develop a policy network capable of selecting actions that maximize cumulative rewards. However, the intrinsic limitation of exploration in DRL restricts its effectiveness in optimizing parameters~\cite{10021988}. Consequently, researchers have proposed approaches that integrate Evolutionary Algorithms (EA) with Reinforcement Learning (RL) to enhance parameter optimization.
The ultimate objective of DRL is to train a policy network capable of selecting actions that maximize the cumulative rewards. However, 
% 梯度优化往往会让策略陷入次优点，阻碍了获得更好的策略
RL with a single policy often exhibits weak exploration capabilities, and gradient-based optimization can easily lead neural networks into local optima, impeding the ability to achieve better performance~\cite{10021988}, To solve the problems, researchers propose integrating EAs with RL to assist parameter optimization.
% It's worth highlighting that this subsection primarily focuses on EA-assisted parameter search. In this context, 
% the related algorithms revolve around RL, with EA playing a complementary role in refining this process. EA cannot optimize independently. 

%we will delve into greater detail in subsection~\ref{sec: parallel}.
%the related works doesn't maintain a complete life-cycle population. Instead, they may periodically construct populations for optimization. For a more comprehensive exploration of life-cycle optimization with RL, we will delve into greater detail in subsection~\ref{sec: parallel}.
% 深度强化学习最终目标是学习一套策略网络能够最大化总体收益。然而DRL单一策略往往探索能力较差，同时基于梯度的学习方式容易使算法学习陷入次优。因此一些学者提出使用EA来辅助策略参数搜索和优化。值得注意的是本节主要是EA辅助的优化，因此没有一个完整的EA种群贯穿学习过程，而RL则会始终被维护。我们会在第X节介绍更多的并行优化与搜索方式。
%EQ引入了一个种群的critic进行提升，并为每个critic配置一个actor，每次种群提升后，对相应的actor进行提升，并基于actor的表现作为critic的fitness进行提升。

% EQ构建了一个Actor和Criic一一对应的种群，其中critic的fitness是相应的actor与环境交互的得分，通过演化critic，再基于演化的critic来使用策略梯度优化actor来实现算法提升。EQ在Inverted Pendulum Task证明了方法的有效性。

EQ~\cite{leite2020reinforcement} employs EAs to replace the conventional Bellman equation for critic optimization. 
Specifically, EQ maintains a critic population, where each critic optimizes a corresponding actor with policy gradients. The fitness of each critic is based on the scores derived from the interactions between its respective actor and the environment. The critics with high fitness are selected as parents and perturbed with Gaussian noise to generate offspring.
%add introducing Gaussian noise for generating offspring. %, which are then employed to generate the subsequent generation by introducing Gaussian noise to their parameters. 
The experiments on the inverted pendulum task demonstrate that using EAs can achieve better guidance capabilities than using conventional Bellman optimization while maintaining comparable performance.
Supe-RL~\cite{marchesini2021genetic} employs GA to auxiliary policy parameter search. At regular intervals, a population of policies is initialized by introducing perturbations to the current RL policy. The population interacts with the environment, and elite individuals soft update their parameters to the RL policy. In off-policy RL, Supe-RL incorporates the elite experiences collected in the genetic evaluation phase into the RL replay buffer. The experiments show that Supe-RL can enhance Rainbow and PPO on navigation tasks and the MuJoCo environment respectively. VFS~\cite{marchesini2023improving} employs the same idea as Supe-RL but differs by the periodic construction of a critic population.
The population is evolved by introducing perturbations of differing scales to the critics. Eventually, the critic with the least deviation from the true value function is chosen to replace the RL critic. The true value function is approximated using unbiased Monte Carlo estimation. VFS demonstrates improvements across various algorithms on MuJoCo tasks, including PPO, DDPG, TD3, and SAC.
% VFS使用演化算法在RL训练过程中每间隔一段时间创建一个Critic的种群，并添加了大小两种强度的扰动，最后选出与true value function之间差距最小的critic来替换掉RL中的critic。其中true value function使用蒙特卡洛方式的无偏估计近似。VFS在MuJoCo上实现了对PPO,DDPG,TD3,SAC等算法的提升。

% VEB-RL则主要专注value-based方法并维护一个value function的种群，并提出了使用负的TD error作为适应度进行种群提升。除此之外，为了避免差的个体产生差的经验影响RL优化，VEB-RL提出了精英交互，只允许精英个体与环境交互来提供多样性经验。VEB-RL在Atari，MinAtar任务中实现了对DQN, Rainbow，SPR等Value Based方法的提升。

% Supe RL的思想与VFS相似，不同点是Supe RL直接在策略网络参数空间进行探索，在RL训练过程中每间隔一段时间，基于当前RL策略网络通过扰动，初始化一个种群个体，并让该种群与环境交互，其中精英个体将自己的参数soft update到RL的策略上。Supe RL在导航任务及MuJoCo中分别实现了对Rainbow和PPO的提升。

\textbf{EA-assisted Action Selection.} 
Action selection runs throughout the entire process of RL improvement and evaluation. It primarily involves optimizing the value function, i.e., computing target values, and interacting with the environment. However, on continuous action tasks, the action space can be vast, making it challenging to determine the optimal actions for optimization and execution, especially in situations involving multimodality or multiple peaks. Traditional RL often employs greedy policies or random sampling from distributions for action selection. However, this approach struggles to accurately capture the best behaviors. Therefore, some works propose the concept of action evolution: initializing a population of actions, evaluating their quality using Q values as fitness, and then selecting the elite action for optimization and interaction. %This evolution process helps identify promising actions for subsequent RL optimization and decision-making.

Qt-Opt~\cite{kalashnikov2018scalable} does not explicitly maintain a policy network, it initializes a population of random actions from the action space. Under the current state, Qt-Opt applies two iterations of the Cross-Entropy Method (CEM)~\cite{rubinstein1999cross} to the population, guided by the Value Function. The best action is then selected for the critic optimization and interaction. Qt-Opt demonstrates its efficiency in real-world robotic visual grasping tasks.
While retaining the advantages of Qt-Opt, CGP~\cite{simmons2019q} reduces computational burdens by introducing a policy to mimic actions sampled by CEM. In CGP, the critic's training uses CEM-derived actions. Concurrently, a policy is trained using behavior cloning or policy gradients. %This policy interacts with the environment, effectively reducing computational overhead than CEM during each decision step. 
CGP validates its efficiency on MuJoCo tasks.
EAS-RL~\cite{ma2022evolutionary} uses a similar idea to CGP, yet it distinguishes itself in two aspects: (1) employing TD3 with the original optimization process and replacing CEM with PSO, and (2) integrating both behavior cloning and policy gradients to optimize the policy. EAS-RL outperforms many ERL-related works on MuJoCo tasks~\cite{khadka2018evolution,pourchot2018cem,bodnar2020proximal,khadka2019collaborative}.
%While retaining the benefits of Qt-Opt, CGP reduces computational burdens by introducing a policy to mimic actions sampled by CEM. In CGP, the critic's training continues to optimize using CEM-derived actions. However, a policy is concurrently trained using behavior cloning or policy gradients. This policy interacts with the environment, effectively reducing computational overhead during each decision step. CGP validates its effectiveness on MuJoCo tasks.
%EAS-RL shares similarities with CGP but differs in three main aspects: Firstly, it replaces CEM with PSO for action selection. Secondly, it combines behavior cloning with policy gradients to optimize the policy. EAS-RL's efficiency is also demonstrated through experiments conducted on MuJoCo tasks.
SAC-CEPO~\cite{shi2021soft} employs a stochastic policy, SAC, instead of a deterministic policy in CGP. SAC-CEPO divides SAC into two parts: a mean network and a deviation network. CEM is employed to select the best mean actions, while the mean network is trained through behavior cloning, and the deviation network is learned using the SAC policy gradient. Besides, the generation of the CEM population is no longer based on random sampling, but on sampling from a normal distribution based on learned mean and deviation networks. SAC-CEPO demonstrates improvements over SAC in MuJoCo tasks. GRAC~\cite{shao2022grac} introduces three mechanisms to improve Double Q-learning. We primarily focus on two mechanisms closely aligned with EAs: CEM Policy Improvement and Max-min Double Q-learning. Similar to previous methods, CEM Policy Improvement employs CEM to search for the optimal actions, the difference is that CEM Policy Improvement uses the Q-value differences between the optimal actions and the actions taken by the RL policy as advantages to increase the probability of the RL policy selecting the optimal actions. Max-min Double Q-learning is proposed to address the underestimation problem of double Q-learning.
Specifically, GRAC obtains the CEM action and RL action based on the next state. Subsequently, it utilizes double Q networks to estimate the minimum Q value for the CEM action and the minimum Q value for the RL action. The higher of the two values is selected as the target value.
GRAC demonstrates significant performance gains in MuJoCo tasks and is also validated on real robots.   

In the offline setting, the same problem is demonstrated to exist: policy gradient improvements tend to get stuck in local optima due to the complex nature of the value function landscape. To solve the problem, OMAR~\cite{OMAR} employs a modified version of CEM to select the most optimal actions and uses behavior cloning to fine-tune the policy. The efficiency of OMAR is validated through experiments on the MPE and D4RL benchmarks.
In the multi-agent setting, COMIX~\cite{COMIX} aims to address the problem of QMIX's inapplicability in continuous action spaces~\cite{QMIX}.
%In the multi-agent setting, a popular method is QMIX~\cite{QMIX}, but it is challenging to apply to continuous action spaces. To solve the problem, COMIX~\cite{COMIX} combines CEM with QMIX.
Within COMIX, QMIX is utilized for value function approximation, followed by CEM for action selection. The experiments demonstrate that COMIX outperforms MADDPG in MPE and MA-MuJoCo tasks.

\textbf{EA-assisted Hyperparameter Optimization.} 
While DRL has shown remarkable prowess across diverse domains, it still suffers from the notorious issue of hyperparameter sensitivity, resulting in a complex and expensive tuning process. Many works solve the problems by incorporating EAs to tune hyperparameters for RL.

OMPAC~\cite{elfwing2018online} employs GA for RL hyperparameter optimization, including the trace-decay rate $\lambda$, discount factor $\gamma$, learning rate $lr$, and temperature $\tau_a$ of softmax action selection. OMPAC trains a population of RL individuals with different hyperparameter configurations and evaluates individuals based on cumulative rewards. After a certain number of iterations, the hyperparameters of non-elite individuals are perturbed by adding Gaussian noise to generate offspring. OMPAC demonstrates significant performance improvements over the basic RL algorithms in Atari and Tetris tasks.
PBT~\cite{PBT} introduces a more generalized optimization framework that can be applied to any neural network training process. Here, we focus solely on the RL aspect. Similar to OMPAC, PBT trains a population of policies with different hyperparameters. After a certain training period, the individuals with high fitness directly replace inferior ones by perturbing their hyperparameters or by resampling hyperparameters from predefined distributions. 
The efficiency of PBT is demonstrated across diverse domains encompassing DM Lab, Atari, and StarCraft II.
SEARL~\cite{franke2021sample} utilizes GA to dynamically adjust the parameters of RL, network architectures (layers, nodes, activation functions), and the learning rates of both actor and critic. 
Similar to PBT, SEARL trains a population of RL individuals and stores experiences generated during the population evaluation phase in a shared replay buffer. Subsequently, SEARL employs GA to add Gaussian noise to network parameters and modify network architecture, and hyperparameters to form a new population. Then SEARL optimizes the population through DRL based on the shared replay buffer.
In four out of five MuJoCo tasks, SEARL outperforms PBT in terms of performance.
GA-DRL~\cite{sehgal2022ga} focuses on adjusting a wider range of hyperparameters. including the discount factor $\gamma$, learning rates $lr$ for both the actor and critic, the soft update coefficient $\tau$, the probability of selecting random actions $\epsilon$, and the variance of the noise $\eta$. These hyperparameters are encoded using an 11-bit binary representation. GA-DRL ranks individuals based on the minimum number of episodes required for the robotic arm to achieve an 85\% success rate. The experiments demonstrate that the algorithms optimized with hyperparameters discovered by GA-DRL can achieve better performance in most of the robotic arm control tasks.
% OMPAC使用GA算法进行RL的超参数寻优，包括：
% the trace-decay rate $\lambda$, discount factor $\gamma$, learning rate $\alpha$, temperature $\tau_a$ of softmax action selection. OMPAC构建了一个不同超参数配置的种群进行并行优化，通过累计奖励评估个体。一定轮数后，基于fitness种群中的非精英个体的超参数通过添加高斯扰动进行超参数探索。随后新的种群进入到下一轮迭代。OMPAC在Atari和Tetris上证明了OMPAC可以显著提升基准算法。
AAC~\cite{grigsby2021towards} dynamically adjusts five hyperparameters: the discount factor $\gamma$, the coefficient for SAC entropy $h$, action duration $k$, and the number of single-step updates $a$ and $c$ for both the actor and critic. 
AAC maintains a population of actor-critic individuals with different hyperparameter configurations and a shared replay buffer. Fitness is defined as the mean of cumulative rewards over multiple episodes. Based on the fitness, the top 20\% best and worst individuals are selected. Superior individuals replace inferior ones with hyperparameter perturbations. Ultimately, AAC demonstrates its efficiency on the DMC benchmark and Industrial benchmark~\cite{hein2016introduction}.
OHT-ES~\cite{tang2020online} employs ES to dynamically adjust RL hyperparameters, including the $n$ parameter in the $n$-step TD and the learning rate ($lr$). OHT-ES samples hyperparameters from distributions maintained by ES and employs these hyperparameters to train several off-policy policies. These policies control agents interacting with the environment for evaluation, and the generated experiences are stored in a shared replay buffer for RL learning. Subsequently, the distributions of hyperparameters are updated. OHT-ES demonstrates significant improvements over the basic algorithm TD3 on the DMC benchmark.

% DEHB~\cite{awad2021dehb} combines DE with Hyperband, maintaining a configuration population for each budget and optimizing them using DE. The incorporation of DE allows the algorithm to leverage previous configurations, instead of solely relying on random sampling like HB. DEHB also employs a parent pool mechanism that allows individuals from lower-budget configurations to participate in the evolution process of higher-budget configurations, facilitating information transfer between budgets. DEHB adjusts 7 hyperparameters of PPO: layer size, batch size, $lr$, $\gamma$, likelihood ratio clipping $\epsilon^{clip}$, and entropy regularization $h$.

% PBT构建了一个具有不同超参数（例如learning rate for A3C）的策略种群，经过一段时间的训练后进行评价，优秀的个体会直接替换掉差的个体，并且会对超参数添加一定的扰动或者从预先定义的分布中重采样来达到超参数探索的目的。PBT在DM Lab， Atari， StarCraft II上都证明了方法的有效性。
% CERL则主要引入GA来调整RL中折扣因子gamma这个超参数，CERL没有使用动态调整的方式，而是初始化了一个配有不同gamma的actor的种群，在训练过程中根据不同actor的表现，来动态的分配资源，从而找到这一组gamma中最好的设定。不同于PBT的是，所有的经验在整个种群中都是共享的。
% SEARL则使用GA动态的调整RL的权重，网络架构（层数，节点数，激活函数）以及Actor和Critic的学习率，并在5个MuJoCo任务中的4个性能超过了PBT。
% GA-DRL则关注更多的超参数进行调整，包括折扣因子，actor和critic的学习率，soft update coefficient \tau, 采取贪心动作的概率，噪音方差。GA-DRL使用11位的二进制对这些超参数进行编码，并判断机械臂达到85%成功率所需要的最小轮数进行个体排序。当GA找到所有合适的超参数后，GA-DRL基于新的超参数训练在大部分机械臂控制任务上都优于DDPG+HER。
% AAC则动态调整五个超参数，折扣因子，SAC熵的系数，动作持续时间，actor与critic的单步更新次数。AAC也使用了经验共享，通过评价种群个体，选择最好和最差的20%的个体，使用好的个体对差的个体进行替换与超参数扰动。最后AAC在DMC上证明了方法的有效性。
% OHT-ES使用ES的方法对超参数n-step以及learning rate进行调整，每次从ES维护的分布中sample N组超参数，随后进行交互，并与环境交互评价基于该超参数训练的个体。基于适应度对分布进行更新，重复该过程。OHT-ES同样在DMC上证明了能对基准算法TD3有显著提升。
% 

\textbf{Others.}
Some works use EAs or the principles of EAs to assist in other aspects of RL, which cannot be categorized into the three categories mentioned above. 

%Evo-Meta RL提出使用演化的思想来搜索损失函数来使得该损失函数可以泛化到不同的环境中。具体来说，Evo-Meta RL提出将损失函数的计算过程转化为计算图，并对其中的Operation nodes进行变异。Evo-Meta RL分为内外两个循环，外循环通过累计奖励来寻找好的计算结构作为父代产生子代，而内循环则基于外循环找到的结构进行梯度优化。Evo-Meta RL最终找到的两个损失函数形成了DQNClipped & DQNReg，并且在Atari和MiniGrid上证明了发现的损失函数优于其他基线DQN,DDQN。

% Evo-Reward构建了一个reward function的种群来进行reward function的搜索，使用PushGP，一个GP算法的变种，来进行种群演化与提升。实验结果表明Evo-Reward在Hungry–Thirsty任务上找到的价值函数要比原始价值函数学习更加高效。
%{Evo-Reward~\cite{Evo-Reward} constructs a population of reward functions and employs PushGP~\cite{PushGP}, a variant of GP, for population evolution to search for more efficient reward functions. Experimental results indicate that Evo-Reward can discover more efficient reward functions than the original reward function on the Hungry–Thirsty task.} 
Evo-Meta RL~\cite{EvoRL} uses EAs to search for the RL loss function capable of generalizing across diverse environments. Specifically, it transforms the computation process of the loss function into a computational graph and introduces mutations to the operation nodes within it. Evo-Meta RL operates in both inner and outer loops; the outer loop identifies the parent computational structures based on cumulative rewards to generate offspring, while the inner loop conducts gradient optimization based on the structures identified in the outer loop. Evo-Meta RL constructs DQNClipped and DQNReg based on the two discovered loss functions and demonstrates their superiority over DQN and DDQN on Atari and MiniGrid tasks.

{GP-MAXQ~\cite{GP-MAXQ} employs GP to explore the hierarchical structure of the hierarchical RL method MAXQ~\cite{MAXQ}. In GP-MAXQ, MAXQ learns policies based on the hierarchies derived from GP, GP explores appropriate hierarchies using MAXQ's outputs. Experimental results on the forgiveness tasks indicate that GP can search for more efficient hierarchical structures.} 
PNS-RL~\cite{liu2021population} aims to improve the exploration capacity of RL. PNS-RL consists of multiple populations, each comprising multiple exploration policies and one guiding policy. Each exploration policy maintains an actor, critic, and replay buffer, and is optimized through policy gradients along with soft updates towards the guiding policy.
Additionally, PNS-RL maintains an archive for selecting the guiding policy. The exploration policy that outperforms the average performance of policies in the archive is added, whereas policies in the archive performing notably worse than the added policy are removed. The most novel individual in the archive is selected as the guiding policy and shared across different populations. 
Novelty is measured based on the distance in the pre-defined behavioral descriptor space between the agent and its nearest $k$ neighbors. The experiments on MuJoCo tasks show that PNS-RL outperforms PBT-TD3, P3S-TD3, CEM-TD3, and others.
Go-Explore~\cite{Go-exp} also focuses on exploration issues. Although it does not utilize RL, it addresses the challenging exploration problem in RL and provides significant inspiration. Go-Explore builds an archive of trajectories, recording trajectories reaching different states. Then it selects the state from the archive that most likely leads to a new state, replicates the trajectory in the environment based on the archive, reaches that state, and starts random exploration from the state. If the newly reached state is not in the archive or reaches an existing state with a more optimal trajectory, the archive is updated. Then the policy is learned directly through imitation learning.
Go-Explore outperforms other RL algorithms and surpasses human performance in challenging exploration problems Montezuma and Pitfall.
G2N~\cite{chang2018genetic} employs a binary GA population to control the activation of hidden neurons in the RL policy network, aiming to enhance the exploration capability. The experiments demonstrate that G2N can improve PPO and A2C on MuJoCo and Atari tasks.
EvoRL~\cite{hallawa2021evo} simulates an evolutionary process, considering that some behaviors are innate and can only be obtained through evolution, while others are learnable. Therefore, EvoRL defines part of the policy's behavior as learnable and employs RL for learning, while another part is represented using behavior trees and can only be evolved through GP.

In addition to the aforementioned works in the single-agent settings, some works focus on improving MARL with EAs.
%严格说，这些工作应该被分到EA辅助RL优化方向，但是由于我们更关注MA的设置，因此在这里介绍。
% Strictly speaking, the following works should be categorized under the direction of EA-assisted RL optimization. However, due to the Multi-Agent setting, we present them here.
% Specifically
ROMANCE~\cite{ROMANCE} focuses on the robust MARL. ROMANCE utilizes the QD algorithm to maintain a population of attackers that sporadically attack some collaborators within the team. The attackers employ policy gradients for individual mutation, incorporating regularization terms for maintaining population diversity and attack frequency. ROMANCE demonstrates its superiority over other robust MARL algorithms on the SMAC benchmark.
Following a similar idea, MA3C~\cite{MA3C} maintains a population of agents to attack the communication channel of cooperative MARL. The individuals are improved by MATD3 and the most novel individual based on policy representation is retained. MA3C demonstrates robust communication across various tasks. 
%including SMAC, Traffic Junction, Gold Panner, and Hallway.
EPC~\cite{EPC} addresses large-scale multi-agent problems by progressively expanding from small to large-scale scenarios in a curriculum-based manner. Larger-scale policies are directly cloned from the policies of the previous scale, but policies that perform well in one scale may not necessarily be suitable for the next larger scale. Thus EPC trains multiple parallel policies at new scales and uses MADDPG as a mutation operator for improvement, ultimately retaining the best-performing policies. In the particle-world environment, EPC demonstrates its ability to efficiently solve large-scale MARL problems.
MAPPER~\cite{MAPPER} focuses on the multi-agent path planning problem in mixed dynamic environments. MAPPER employs EAs to enhance the stability of RL training by maintaining superior individuals and eliminating inferior ones. 
The mutation operator uses RL to improve the individuals, Each policy can be replaced by another policy with a certain probability that decreases as the score increases. MAPPER uses A2C instead of MARL algorithms.

A related line of work to Evo-Meta RL uses hardware acceleration to make the inner loop training far faster and more tractable \cite{lu2022discovered}. By doing this, 'Learned Policy Optimisation' \cite{lu2022discovered} and 'Temporally-Aware Learned Policy Optimisation' \cite{jackson2024discovering} discover alternative on-policy RL algorithms that significantly outperform PPO on evaluation tasks. Hardware-accelerated evolutionary meta-RL has also been applied to evolve reward functions \cite{sapora2024evil}, offline datasets \cite{lupu2024behaviour}, and adversarial environments \cite{lu2023adversarial}.

\textbf{Challenges and Future Directions}: 
The utilization of EAs to assist RL through parameter search, action selection, hyperparameter tuning, and other aspects has demonstrated the potential to further improve the performance of RL. %These approaches have been validated in various sequential decision-making tasks, including navigation, locomotion, and real-world robot control.
However, utilizing EAs for RL optimization still faces several challenges:
1) Researchers require domain knowledge to define the individual form, fitness function, variation, and selection operators.
2) EAs often require a substantial number of evolutionary iterations. This takes additional computational costs and may introduce extra sample costs.
3) EAs introduce additional hyperparameters, complicating the application and tuning.
4) The works involving EA-assisted RL primarily demonstrate their efficiency in SDP. However, there is a lack of validation and systematic analysis in other problems, e.g., CTOP.
%Addressing these challenges may require further research and algorithmic improvement to enhance the usability and efficiency of the approach.
In the following, we outline several prospective research directions for the future:
1) Establishing an automated configuration mechanism for EAs to enhance their usability.
2) Enhancing the efficiency of EAs, such as constructing a more sample-efficient population evaluation method.
3) Introducing EAs that are more efficient, robust, and less sensitive to hyperparameters.
4) The role of EAs in RL, beyond the aforementioned branches, remains open for further exploration. 5) There is a need for deeper investigation into the potential of EA-assisted RL in various other optimization problems.
%This could involve more efficient, robust, and hyperparameter-insensitive evolutionary algorithms, among others.

\begin{table*}[htbp]
  \renewcommand{\arraystretch}{1.4}
  \centering
  \caption{RL-assisted Optimization of EA. %We employ abbreviations for all tasks, with the majority of them being introduced in Subsection~\ref{problem defined}. Tasks that have not been included are as follows: symbolic regression problem (SRP), black box optimization problem (BBOP); white box optimization problem (WBOP).
  % 我们简化了任务名称，其中SRP是Symbolic Regression Problem， COP是Combinatorial Optimization Problem， HOP是H-IFF optimization problem，TSP是Travelling salesman problem， VRP是Vehicle Routing problem，MOP是多目标优化问题， VRP是 Vehicle Routing problem ， TOP是Trajectory Optimization Problems ， SP是 Scheduling Problem ， MP是Multimodal Problem，WBP是White-Box Optimization Problems, MOSP(MO Scheduling) 
  }
  \begin{tabular}{l | lllll}
    \toprule
    \textbf{Class} & \textbf{Algorithm} &  \textbf{Problems} & \textbf{EA} & \textbf{RL}  & \textbf{Detail} \\
       \midrule
    \multirow{2}{*}{\makecell[l]{Population \\ Initialization}} 
    & {NGGP~\cite{NGGP}} & \makecell[l]{SRP }  & GP & PG  & Provide individuals with PG \\   
        \cline{2-6}
    &{RL-guided GA~\cite{RL_guided_EA}}   &  \makecell[l]{%nuclear fuelassembly
   COP} & GA & PPO    & Provide individuals with PG\\
           \cline{2-6}
    & DeepACO~\cite{DeepACO} &  \makecell[l]{%nuclear fuelassembly
   COP} & ACO & REINFORCE   & Provide individuals with PG\\
    \hline
    \multirow{3}{*}{\makecell[l]{Population \\ Evaluation}} & {SC~\cite{wang2022surrogate}} & SDP &EA & DDPG  & Use Critic and Replay Buffer \\
    \cline{2-6}
    \multirow{1}{*} & {PGPS~\cite{kim2020pgps}} & SDP & CEM & TD3 & Use Critic and Replay Buffer \\
    \cline{2-6}
    \multirow{1}{*} & {ERL-Re$^2$~\cite{Re2}} & SDP & EA & DDPG, TD3  &  $H$-step Bootstrap\\
    \hline

  \multirow{3}{*}{\makecell[l]{Variation \\ Operator}}
    &{GPO~\cite{GangwaniP18GPO}} & SDP & GA & PPO or A2C & Apply PG for mutation \\
    \cline{2-6}
    \multirow{1}{*} &  {CEM-RL~\cite{pourchot2018cem}} & SDP & CEM & TD3 & Apply PG for mutation \\
    \cline{2-6}
    \multirow{1}{*} & {CEM-ACER~\cite{tang2021guiding}} & SDP & CEM & ACER  & Apply PG for mutation \\
    \cline{2-6}
    \multirow{1}{*} & {PBRL~\cite{pretorius2021population}} & SDP & GA & DDPG  & Apply PG for mutation \\
    \cline{2-6}
    \multirow{1}{*} & {NS-RL~\cite{Shi2020ffficient}} & SDP & NS & DDPG  &  Apply PG for mutation \\
    \cline{2-6}
    \multirow{1}{*} & {DEPRL~\cite{liu2021diversity}} & SDP & CEM & TD3  & Apply PG for mutation\\
    \cline{2-6}
    \multirow{1}{*} & {QD-RL~\cite{QDRL}} & MOOP \& SDP & Map-Elites-like & TD3  & Guide search with QD Critics\\
    \cline{2-6}
    \multirow{1}{*} & {PGA-ME~\cite{nilsson2021policy}} & MOOP \& SDP & Map-Elites & TD3  & Half with PG and Half with EA  \\
    \cline{2-6}
    \multirow{1}{*} & {GAC QD-RL~\cite{GAC-QD-RL}} & MOOP \& SDP & Map-Elites & SAC \& DroQ  & Half with PG and Half with EA  \\
    % \cline{2-6}
    % \multirow{1}{*} & {EDO-CS} & \makecell[l]{Maze Task \& \\ MO MuJoCo} & Map-Elites & TD3 & search with policy gradients for half of the population  \\
    \cline{2-6}
    \multirow{1}{*} & {CMA-MEGA~\cite{tjanaka2022approximating}} & MOOP \& SDP & Map-Elites &  TD3 & Optimize the fitness with PG \\
    \cline{2-6}
    \multirow{1}{*} & {CCQD~\cite{anonymous2024sampleefficient}} & MOOP \& SDP & Map-Elites &  TD3 & \makecell[l]{Half with PG and Half with EA  \& \\ Construct the shared representations}  \\
    \cline{2-6}
    \multirow{1}{*} & {RefQD~\cite{wang2024quality}} & MOOP \& SDP & Map-Elites &  TD3 & \makecell[l]{Half with PG and Half with EA  \& \\ Construct the shared representations}  \\
    \cline{2-6}
    & {Wuji~\cite{wuji}} & \makecell[l]{MOOP}  & MOEA & A2C  &  Apply PG to offspring \\  
    \hline
   \multirow{4}{*}{\makecell[l]{Dynamic \\ Operator \\ Selection}} &  {RL-GA(a)~\cite{DBLP:conf/gecco/PettingerE02}} & COP  & GA & Q-learning&  Variation operators, parent types\\
    \cline{2-6}
    \multirow{1}{*} & {RLEP~\cite{RLEP}} & CTOP  &  ES  & Q-learning & Variation operators \\
    \cline{2-6}
    \multirow{1}{*} & {EA+RL~\cite{EA+RL}} & COP  &  EP  & Q-learning & Fitness Function \\
    \cline{2-6}
    \multirow{1}{*} & {EA+RL(O)~\cite{EA+RL(O)}} & COP  &  GA  & Q-learning & Variation operators \\
    \cline{2-6}
    \multirow{1}{*} & {RL-GA(b)~\cite{DBLP:journals/swevo/SongWYWXC23}} & COP & GA & Q-learning & Variation operators\\
    \cline{2-6}
    \multirow{1}{*} & {GSF~\cite{yi2022automated}} & COP &  GA  & DQN, PPO &  All operators during all stages \\
    \cline{2-6}
    \multirow{1}{*} & {MARLwCMA~\cite{DBLP:journals/access/SallamECR20}} & CTOP  & \makecell[l]{DE \\ CMA-ES}  & Q-learning & Variation operators\\
        \cline{2-6}
    \multirow{1}{*} & {MPSORL~\cite{meng2023multi}} & CTOP  &  PSO  & Q-learning &  Variation operators\\
    \cline{2-6}
    \multirow{1}{*} & {DEDQN ~\cite{DEDQN}} & CTOP &  DE & DQN &  Variation operators \\
    \cline{2-6}
    \multirow{1}{*} & {DE-DDQN~\cite{sharma2019deep}} & CTOP  & DE  & DDQN &  Variation operators \\
    \cline{2-6}
     \multirow{1}{*} & {RL-CORCO~\cite{RL-CORCO}} & CTOP &  DE  & Q-learning &  Variation operators\\
    \cline{2-6}
    \multirow{1}{*} & {RL-HDE~\cite{RL-HDE}} & CTOP &  DE  & Q-learning &  Variation operators, parameters\\
    \cline{2-6}
    \multirow{1}{*} & {DE-RLFR~\cite{DBLP:journals/swevo/LiSYSQ19}} & MOOP &  MODE  & Q-learning &  Variation operators\\
    \cline{2-6}
    \multirow{1}{*} & {LRMODE ~\cite{huang2020afitness}} & MOOP &  MODE & Q-learning &  Variation operators \\
    \cline{2-6}
    \multirow{1}{*} & {MOEA/D-DQN~\cite{tian2022deep}} & MOOP &  MOEAs  & DQN &  Variation operators \\
    \cline{2-6}
    \multirow{1}{*} & AMODE-DRL~\cite{li2023scheduling} & MOSP &  MODE  & \makecell[l]{DDQN,DDPG} &  Variation operators, parameter \\
    \hline    
    \multirow{3}{*}{\makecell[l]{Dynamic \\ 
    Hyperparameter\\
    Configuration}} & AGA~\cite{DBLP:conf/esoa/EibenHKS06} & MMOP & GA &  \makecell[l]{Q-learning\\, SARSA}& \makecell[l]{Crossover rate, mutation rate, \\tournament size, population
size}  \\
    \cline{2-6}
    & LTO~\cite{LTO}  & BBOP & CMA-ES & GPS~\cite{GPS}    &  Mutation step-size parameter\\
    \cline{2-6}
    
    & RL-DAC~\cite{DBLP:conf/ecai/BiedenkappBEHL20}  & WBOP & - &  \makecell[l]{Q-learning,\\ DDQN}    &   Formalize DAC as MDP \\
    \cline{2-6}
    & REM~\cite{zhang2022variational}  & CTOP  & DE & VPG    &  Scale factor, crossover rate\\
    \cline{2-6}
    & \makecell[l]{Q-LSHADE \& \\ 
     DQ-HSES~\cite{DBLP:journals/cim/ZhangSBZX23}} & CTOP & \makecell[l]{LSHADE \\\, HSES} &  \makecell[l]{Q-learning \\     , DQL}    &  The switching time\\
    \cline{2-6}
    
    \multirow{1}{*} & {MADAC~\cite{DBLP:conf/nips/0001XYL0Z022}} & MOOP & MOEA/D & VDN   & Hyperparameter, operators\\
    \cline{2-6}
    & qlDE~\cite{qlDE}  & CTOP &          
    DE &  Q-learning   &  Scale factor, crossover rate\\
    \cline{2-6}
    & RLDE~\cite{hu2021reinforcement}  & CTOP & DE &  Q-learning   &  Scale factor\\
    % \multirow{1}{*} & {FIDI-DDPG} & ARS & DDPG    & \\
    % \cline{2-5}
    \hline
    
    \multirow{2}{*}{\makecell[l]{Others}} 
    & {RGP~\cite{RGP}} & \makecell[l]{SDP}  & GP & Q-learning  &  Improve  efficiency \\  
    \cline{2-6}
    & {GNP-RL~\cite{GNP-RL}}  &  \makecell[l]{SDP} & GNP & Q-learning & Improve efficiency\\
    \cline{2-6}  
    & {Grad-CEM~\cite{Grad-CEM}} & \makecell[l]{SDP}  & CEM & SGD  &  Improve CEM efficiency \\  
            \cline{2-6}
    & {LOOP~\cite{LOOP}}  &  \makecell[l]{SDP} & CEM & SAC & Enhancing planning\\
        \cline{2-6} 
    & {TD-MPC~\cite{TD-MPC}}  &  \makecell[l]{SDP} & CEM & DDPG &  Enhancing planning\\
    \bottomrule
  \end{tabular}
\end{table*}

\section{RL-assisted Optimization of EA}
% 本节主要提供一个RL辅助EA的全面概括。进化算法的优化流程可以概括为，算法配置，种群初始化，父代选择（选择算子），变异，种群评估，子代选择。

%我们根据RL所影响的EA阶段将相关工作分为了五个类别，主要包括RL种群初始化，RL辅助种群评估，基于RL的选择算子，基于RL的变异算子，基于RL的EA算法动态配置五个方面。
% \begin{figure}[t]
% \centering
% \includegraphics[width=1.0\linewidth]{Survey/EA_master.PDF}
% %\vspace{-0.2cm}
% \caption{Schematic for RL-Assisted Optimization of EA.}
% \label{Figure: RL-Assisted Optimization of EA}
% %\vspace{-0.2cm}
% \end{figure}
In this section, we move on to the opposite side and present a comprehensive overview of RL-assisted EA. The optimization schematic is illustrated in Figure~\ref{Figure: RL-Assisted Optimization of EA}. In this optimization process, the related algorithms revolve around EAs, with RL playing a supporting role in refining this process. RL cannot optimize independently. 
We categorize the related works into six branches based on the impact of RL on different stages of EAs. These branches include RL-assisted population initialization, RL-assisted Population Evaluation, RL-assisted Variation Operators, RL-assisted Operator Selection, RL-assisted Dynamic Hyperparameter Configuration, and Others. Each branch focuses on utilizing RL to improve specific stages of EAs, with a dedicated focus on solving distinct problems.
Specifically, Population Evaluation, Variation
Operators and Others primarily focus on SDP.
Among these, Population Evaluation primarily focuses on leveraging the discriminative capability of RL value functions to evaluate individuals, aiming to enhance the sample efficiency of EAs. Variation Operator concentrates on using RL value functions to provide directional guidance for more efficient variation. Others primarily aim to enhance the evolutionary efficiency with RL or improve the accuracy of fitness for planning using the RL value function.
Population Initialization, Dynamic Operator Selection, and Dynamic Hyperparameter Configuration primarily focus on CTOP, COP, MOOP, etc.
Among these, Population Initialization primarily aims to leverage RL's learning capability to provide initial solutions for the population. %replacing heuristic solutions, and improving the optimization efficiency of EAs.
Dynamic Operator Selection focuses on the issue of operator sensitivity in EAs, i.e., no single operator can perfectly solve all problems. The goal is to use RL to achieve automated operator selection, enhancing the robustness of EAs.
Dynamic hyperparameter configuration primarily focuses on utilizing RL to automatically configure the algorithm's hyperparameters in EAs.
Below, we introduce each branch of this direction along with the related algorithms.

\textbf{Population Initialization.} 
Population initialization is the initial step for all EAs, where solutions are randomly or heuristically provided as initial candidates. 
A well-designed population initialization can significantly enhance the search efficiency of EAs, while poor initialization can hinder the algorithms from finding superior solutions. Some works have shown that leveraging known high-quality initial solutions can greatly improve algorithm performance~\cite{lobo2015inferring}. %Besides, the approach that employs policy gradients for automatic initialization can also yield stable and controlled solutions in limited-size populations. 
Consequently, several works propose to leverage RL to improve the quality of the initial population.

To address the Symbolic Regression Problem (SRP), NGGP~\cite{NGGP} employs a policy-gradient-guided sequence generator for population initialization. Subsequently, a certain number of GP iterations are conducted, and the top $E$ individuals from the population are combined with the initial population to train the sequence generator. Subsequently, the next iteration process begins.
Experiments demonstrate that NGGP outperforms previous algorithms on the SPR benchmark.
Similarly, RL-guided GA~\cite{RL_guided_EA} employs PPO to master and match problem rules and constraints. Then the trained policies are used for population initialization. In each generation, RL learns to get new policies, which are added to the population. Experimental results demonstrate that RL efficiently learns the rules to generate candidate solutions for nuclear fuel assembly optimization problems.
%不同于上述直接使用RL提供解的方案，DeepACO~\cite{DeepACO}则使用基于REINFORCE训练的图神经网络初始化ACO的启发式度量，常规ACO中的启发式度量通常是基于专家知识预先定义的。而DeepACO使用RL选择每个节点的概率来初始化该度量，一方面避免了专家知识的引入，一方面加速了求解效率。DeepACO在九个组合优化问题上例如TSP，SOP等带来了显著的性能增益。

Differing from the aforementioned approaches that directly employ RL to provide initial solutions, DeepACO~\cite{DeepACO} utilizes a graph neural network trained with REINFORCE to initialize the heuristic measure of Ant Colony Optimization (ACO). 
In conventional ACO, the heuristic measure is typically predefined based on expert knowledge. 
DeepACO uses the probability of RL selecting each node to initialize the heuristic measure, avoiding the introduction of expert knowledge and simultaneously accelerating the solving efficiency. DeepACO brings significant performance gains on nine combinatorial optimization problems, such as TSP, SOP, and others.

\textbf{RL-assisted Population Evaluation.} The population evaluation is a crucial step in obtaining fitness. However, when applying EAs to address sample-cost-sensitive problems, such as robot control, the evaluation process requires applying each solution in the population to the problem to obtain fitness. This typically incurs a substantial sample cost.
%Evaluating the population is an essential step to obtaining the fitness ranking for evolution. However, When we use EAs to solve sample-cost-sensitive problems, such as robot control problem, the evaluation process requires each individual in the population to be applied to the problem for obtaining fitness, which typically incurs a significant sample cost.
% While traditional methods discard evaluation experiences, leading to sample inefficiency. 
To solve the problem, several works propose using RL value functions to evaluate the fitness of EA individuals, thereby improving sample efficiency.

SC~\cite{wang2022surrogate} utilizes the expected Q values estimated using RL critics and experiences as the fitness surrogate to evaluate the population. It proposes two approaches: 1) Probabilistic evaluation using the surrogate. 2) Selecting a population twice the size of the original and filtering half using the surrogate. %Experiments show that both approaches can lead to performance improvement on MuJoCo tasks.
PGPS~\cite{kim2020pgps} adopts the second approach of SC for population evaluation. 
ERL-Re$^2$~\cite{Re2} introduces $H$-step bootstrap for population evaluation. Each interacts with the environment for $H$ steps, and then the value function is used to estimate the Q value of ($H+1$)th state, which are then summed with the extrinsic rewards in the form of cumulative discount to serve as the fitness. ERL-Re$^2$ probabilistically applies $H$-step bootstrap. All the above methods have shown significant improvements in the sample efficiency. Strictly speaking, the aforementioned works should be classified as instances of synergistic optimization of EA and RL. 
But in this context, we focus on the population evaluation using RL. We provide a detailed explanation of these methods in section~\ref{sec: parallel}.

\textbf{RL-assisted Variation Operators.} Traditional variation operators are typically gradient-free and rely on random search, which requires extensive exploration to find feasible solutions, resulting in low exploration efficiency. To improve efficiency, some works incorporate the policy gradient guidance into EAs to assist with variation operations. These works can be categorized into two main classes: Single-Objective Optimization and Quality-Diversity Optimization.

%在单目标优化中：算法希望通过EA找到能够最大化收益的解。GPO使用ppo或A2C来将策略梯度应用到种群个体上，辅助变异操作。
%而CEM-RL在CEM中引入了TD3的critic，随机从种群选择一半的个体用于优化TD3 Critic，随后将critic提供的策略梯度注入到这些个体中进行提升，而另一半个体通过添加高斯噪音进行策略搜索。融入TD3 critic的CEM的性能在MuJoCo的三个任务中超越了CEM和TD3的性能。CEM-ACER则与CEM-RL相同但是将TD3替换为了ACER。

\emph{(1) Single-Objective Optimization.} This category of algorithms aims to find solutions that maximize a single objective.
GPO~\cite{GangwaniP18GPO} devises gradient-based crossover and mutation by policy distillation and policy gradient algorithm.
CEM-RL~\cite{pourchot2018cem} employs the TD3 critic to guide the CEM mutation. In CEM-RL, half of the population is randomly selected and used to optimize the TD3 critic. The policy gradient from the critic is then injected into these selected individuals for mutation. The other half of the population conducts policy search by adding Gaussian noise. CEM-RL outperforms CEM and TD3 in three tasks of OpenAI MuJoCo.
Another algorithm in this category is CEM-ACER~\cite{tang2021guiding}, which follows a similar framework as CEM-RL but replaces TD3 with ACER.
% PBRL与CEM-RL架构相似，除了替换CEM为GA，TD3为DDPG外，PBRL让种群中变异个体分别与环境交互一定步数，并且在从replay buffer采样进行梯度更新时，以一定比例混合了该个体当前收集的经验以及其他个体生成的经验。
% 除此之外，PBRL提供了自动超参数调整版本 Hyper-parameter tuning PBRL，通过为种群中每个个体添加一个相应的超参数，并在变异时添加随机扰动以该使用该超参数优化前后的提升作为适应度进行超参数搜索。
PBRL~\cite{pretorius2021population} is similar to CEM-RL but with a key difference of replacing CEM with a GA while incorporating DDPG for mutations. Specifically, PBRL allows individuals to interact with the environment for some steps and performs individual gradient optimization using a blend of experiences from the current individual and others.
Furthermore, PBRL presents an automated hyperparameter tuning version called Hyperparameter tuning PBRL. In this version, each individual in the population is associated with a corresponding hyperparameter. Gaussian perturbations are added for hyperparameter mutation. The improvement using the current hyperparameter is used as the fitness.
PBRL and Hyperparameter tuning PBRL demonstrate superior performance over GPO and DDPG in four MuJoCo tasks.
Instead of maximizing performance, NS-RL~\cite{Shi2020ffficient} integrates DDPG to improve the exploration capabilities. In NS-RL, the fitness is defined as the L2 distance between a policy and the k-nearest policies to it within the behavior characterization space. The most novel individual is selected as the elite, while the less novel individuals are improved by minimizing the difference in their novelty compared to the elite. Furthermore, NS-RL takes the goal as the additional input of the critic to enhance the generalization. 
NS-RL demonstrates its efficiency on maze tasks.
DEPRL~\cite{liu2021diversity} considers the diversity of policies, but the ultimate objective remains to maximize the rewards. 
DEPRL follows CEM-RL and employs Maximum Mean Discrepancy (MMD) to measure the distance among policies as the diversity metric. By concurrently maximizing rewards and MMD with gradient optimization and taking rewards and MMD as the fitness metric, DEPRL improves the diversity and exploration capabilities of the population. 
DEPRL outperforms CEM-RL in some MuJoCo tasks.

\emph{(2) Quality-Diversity Optimization.} This category of algorithms diverges from Single-Objective Optimization, considering two ultimate objectives: solution quality and solution diversity.
QD-RL~\cite{QDRL} introduces two TD3 critics into the QD framework, one for quality and the other for diversity. 
The calculation method of diversity is the same as NS-RL. QD-RL maintains an archive to save all past policies. At the start of each iteration,
QD-RL selects individuals from the diversity-quality Pareto front constructed from the archive. Half of the selected individuals are optimized using the quality critic, and the other half are optimized using the diversity critic. 
Finally, the offspring are evaluated and inserted into the archive.
QD-RL outperforms TD3 and other QD methods in exploration and deceptive reward maze tasks.
PGA-ME~\cite{nilsson2021policy} follows a similar process but differs in that half of the population is mutated using the original operator of Map-Elites, while the other half employs the TD3 critic for mutation. PGA-ME outperforms QD-RL and other QD methods in QDMuJoCo tasks. 
{Furthermore, another study~\cite{DBLP:journals/telo/FlageatCC23} highlights the decisive role of the policy-gradient variation operator in PGA-ME, particularly in the early optimization stages. Moreover, the study shows that  PGA-ME demonstrates robust performance in both deterministic and stochastic environments, with solutions found in stochastic settings proving highly reproducible.
GAC QD-RL~\cite{GAC-QD-RL} proposes a general framework by integrating SAC and DroQ~\cite{hiraoka2021dropout} into PGA-ME. It reveals three insights: enhancing the update-to-data ratio (UTD), which represents the frequency of updating the critic when new transitions are collected, leads to improved performance; Unlike DroQ, the diverse data distribution in QD-RL makes it difficult for the critic to overfit, hence the regularization term in DroQ is not necessary; PGA-ME (SAC) is more efficient than PGA-ME (TD3) in tasks with low-dimensional behavioral descriptor space, while its performance is inferior in tasks with high-dimensional behavioral descriptor space.}
CMA-MEGA~\cite{tjanaka2022approximating} maintains a distinct policy for training instead of sampling from the archive. It utilizes the features of the behavioral description space as diversity metrics. CMA-MEGA estimates the gradients of the diversity metric through OpenAI-ES, the gradients of the quality through TD3, and generates gradient coefficients for the population via CMA-ES to regulate the degree of optimization towards different objectives. Following this, it evaluates individuals within the population optimized with varying coefficients, incorporates them into the archive, prioritizes individuals based on whether new grids are filled in the archive or if grids are elevated, and then updates the CMA-ES. Lastly, optimize the TD3 critic. However, CMA-MEGA performs worse than PGA-ME on QDMuJoCo.
CCQD~\cite{anonymous2024sampleefficient} draws inspiration from ERL-Re$^2$~\cite{Re2} by similarly decomposing the policy into shared representations and independent policy representations to facilitate knowledge sharing.
Unlike ERL-Re$^2$, CCQD maintains multiple shared state representations to construct different knowledge spaces, enhancing the algorithm through a cooperative evolution approach.
Each policy requires a unique combination of state representations and policy representations. Based on different shared representations, the behaviors of policies may vary significantly.
Policies discovered during the learning process are inserted into the archive. CCQD outperforms previous QD algorithms, reaching a new state-of-the-art performance on QDMuJoCo.
RefQD~\cite{wang2024quality} also employs the shared state representation and attempts to address the mismatch between old and new policies introduced by the shared state representation in Map-Elites. It periodically re-evaluates the archive and weakens the elitist mechanism of QD by maintaining more decision parts in each archive cell. RefQD outperforms PGA-ME under limited resources.
Wuji~\cite{wuji} uses A2C to further enhance the offspring produced by crossover and mutation in MOEA, which can be seen as an additional variation operator. Wuji demonstrates superior performance in game bug detection.

% CCQD使用了借鉴了ERL-Re$^2$的思想，将共享状态表征引入到了多目标优化当中，将策略拆解为共享表征与独立策略表征。
% 不同于ERL-Re$^2$，CCQD维护了多个共享状态表征来构建不同的知识空间，通过协同演化的思想进行算法提升。将学习过程中发现的解插入到map elites中。CCQD在QDMuJoCo上优于以往QD算法达到了新的性能SOTA。
%NS-RL则不在最大化环境反馈的收益，而使用DDPG进一步提升NS的探索能力。其中优化目标被定义为在行为刻画空间中策略与与距离其最近的k个策略的L2距离。种群中最新颖的个体被作为精英。种群中最差的个体通过使用RL策略梯度来最小化其与种群中最新颖个体新颖度的差距来提升个体新颖度。同时由于Critic每次优化个体不同，因此引入了goal到critic中来提升判别器的重用能力。

%质量多样性：这类算法不在关注单一目标，而是关注两个目标，一个是解的质量，另一个是解的多样性。
%DEPRL在CEM-RL框架下通过使用 maximum mean discrepancy（MMD）来衡量当前策略与前一代策略的距离，在最大化收益的同时也最大化MMD, 来提升种群多样性与探索能力。
%QD-RL引入了两个TD3 critic分别拟合quality和diversity用于提升map elite。其中多样性的计算方式与NS-RL相同。具体来说，QD-RL维护了一个帕累托前沿或者一个archieve存储多样化的解。在每次迭代开始时，首先选择一定的个体形成种群，QD-RL各选择的一半的个体用于优化quality Critic和diversity Critic，随后使用策略梯度分别注入到相应的个体上，替代变异操作，最后对后代进行评价，如果优于archeive中对应的个体，则替换否则丢弃。
%PGA-ME使用了类似的过程，不同点是其中一半的种群个体使用GA进行提升，另一半使用基于外在奖励训练得到的TD3进行优化与提升。
%CMA-MEGA也引入了archieve，不同点是维护一个单独的策略用于训练，而不在从archieve中采样，并将描述行为空间的特征作为多样化衡量进行优化，CMA-MEGA通过OpenAI-ES来估计衡量的梯度，使用TD3衡量收益的梯度，并使用CMA-ES来生成一个种群的梯度系数用于控制向不同目标优化的程度，随后评价基于不同系数优化后的种群个体并尝试插入到archieve中，archieve是否有新的grid被填充以及是否有grid被提升来对个体进行排序，并更新CMA-ES，最后使用off policy的策略训练TD3。

\textbf{RL-assisted Dynamic Algorithm Configuration.}
The utilization of EAs faces several significant challenges during both the configuration and application stages. Firstly, no single EA operator can efficiently solve all problems, leading to the need for a selection of EA operators based on problem characteristics and expert insights. Secondly, EAs are highly sensitive to hyperparameters, demanding meticulous adjustments. Even slight changes can lead to significant performance differences.
To solve the problems, many works improve the usability and robustness of EAs by dynamically selecting the operators and tuning hyperparameter configuration, which is commonly referred to as Dynamic Algorithms Configuration (DAC). In this context, our focus lies primarily on how RL can assist in the DAC process. We categorize these works into two major domains: Dynamic Operator Selection and Dynamic Hyperparameter Configuration.

\textbf{Dynamic Operator Selection.} The algorithms discussed here primarily focus on COP, MOP, MOOP, and CTOP.
% RLEP使用Q-learning来为EP动态的选择四个变异算子。RLEP根据后代相较于父代的提升定义reward直接拟合几个变异算子的累计折扣奖励。RLEP在functional
% optimization problems上优于或者相同于四个基础mutation算子。
RL-GA(a)~\cite{DBLP:conf/gecco/PettingerE02} employs $Q(\lambda)$ to enhance GA by dynamic operator and parent type selections. 
The population itself forms the states and the rewards are defined as the improvement of the offspring compared to the parents.
%Specifically, the current population constructs the states as the input of RL. The reward signals of RL are defined as the improvement of the offspring compared to the parents.
RL selects crossover and mutation operators along with specifying the parent types to which these operators should be applied. 
%In this context, the top 10\% of the population is classified as F, while the remaining are designated as U. For the crossover operator, parents are drawn from four possible types: \{FU, FF, UU, UF\}. Similarly, for the mutation operator, parents are chosen from \{F, U\}. 
RL-GA(a) outperforms GA on the Traveling Salesman Problem.
RLEP~\cite{RLEP} employs Q-learning to dynamically select four mutation operators for EP. RLEP defines the rewards as the improvements of offspring over parents and directly approximate expected returns for the four mutation operators. RLEP outperforms or performs equivalently to the four basic mutation operators on functional optimization problems.
EA+RL~\cite{EA+RL} employs RL to dynamically select the fitness function to enhance the optimization efficiency of GA under the target fitness. The rewards are defined as the performance differences between the best individuals under the target fitness at sequential generations. The states are constructed based on the fitness values of the population. EA+RL demonstrates improvements over ES in the Royal Roads problem and H-IFF optimization problem.
EA+RL(O)~\cite{EA+RL(O)} dynamically employs Q-learning to select crossover and mutation operators for the next generation. 
Similarly to EA+RL, the rewards are defined as the differences in performance between the best individual and its predecessor in the previous generation. 
The RL states are tailored for different tasks.
The efficiency of EA+RL(O) is validated on the H-IFF optimization problem and the Traveling Salesman Problem.
RL-GA(b)~\cite{DBLP:journals/swevo/SongWYWXC23} employs RL to enhance GA in the electromagnetic detection satellite scheduling problems. The definition of rewards is consistent with that of RL-GA(a), and the states are formulated by considering the fitness improvement and the original fitness values.

GSF~\cite{yi2022automated} employs DQN and PPO to dynamically select appropriate combinations of algorithmic components (i.e., evolution operators) during different optimization stages in the capacitated vehicle routing problem with time window (CVRPTW)~\cite{CVRPTW}.
GSF encodes essential information required to solve the CVRPTW problem as states, including fitness, the number of vehicles, and capacity.
The rewards are determined by the performance improvement of the current population compared to the initial population.
GSF demonstrates the efficiency of both PPO-GSF and DQN-GSF in the CVRPTW.
MARLwCMA~\cite{DBLP:journals/access/SallamECR20} proposes a framework that combines multiple optimization algorithms, including multi-operator DE and CMA-ES. In this framework, multi-operator DE dynamically selects mutation operators with the assistance of RL.
The rewards are defined as the cumulative performance improvement of offspring generated using the selected operators compared to their parents.
The states contain two variables designed to reflect population diversity and quality. 
MARLwCMA outperforms other EAs on multiple CEC benchmarks.
In MPSORL~\cite{meng2023multi}, the action space consists of four strategies, while states are divided unevenly into five grades based on fitness values.
Subsequently, MPSORL selects the optimal strategy for each particle. To update the Q-table, a reward of 1 is returned if the particle improves; otherwise,  a reward of 0 is returned.
MPSORL demonstrates superior performance compared to other PSO algorithms on the CEC benchmark.
% MPSORL employs Q-learning to enable each particle in Particle Swarm Optimization (PSO) to select its mutation operator. 
% MPSORL的动作为四类策略，包括A,B,C和D，状态按照fitness大小被不均匀的划分为了五类。随后MPSORL选择最佳策略应用到每个粒子。如果粒子有提升，则奖励为1，否则为0。
% MPSORL在CEC benchmark上优于其他PSO算法。
DEDQN~\cite{DEDQN} utilizes DQN to dynamically select from three mutation operators in DE, primarily divided into two stages. In the first stage, DQN is trained, where states are constructed based on the information from fitness landscape theory~\cite{wright1932roles}, including fitness distance correlation and ruggedness of information entropy. The rewards are constructed from the live algebraic and the individual evolutionary efficiency. In the second stage, the trained DQN selects mutation operators to improve DE. DEDQN demonstrates its superiority over five well-known DE variants in the CEC2017 benchmark.
DE-DDQN~\cite{sharma2019deep} utilizes DDQN to dynamically select mutation operators for each parent in DE. The RL state space comprises a 99-feature vector to capture the DE's current state. 
The reward function takes three forms: R1, reflecting the fitness differences between offspring and parents; R2, assigning a higher reward for improvements over the best solution compared to improvements over the parents; and R3, concurrently maximizing offspring's fitness differences while minimizing the gap between offspring and the optimal solution. DE-DDQN demonstrates superior performance over other DE methods and dynamic operator selection methods in the CEC benchmark.
%DE-DDQN使用DDQN为DE中每个父代选择变异算子。RL的状态是包含99个特征的向量，用于描述DE的状态。奖励函数则有三种，R1定义为the fitness difference of offspring from parent. R2则assigns a higher reward to an improvement over the best so far solution than to an improvement over the parents. R3 maximize the fitness difference offspring and minimize fitness difference between offspring and optimal solution. DE-DDQN在CEC benchmark上证明了其优于其他DE方法和动态算子选择方法。
%DE-DDQN integrates DDQN into DE and selects the mutation operator for each parent. These three algorithms all demonstrate that RL can enhance EAs through the dynamic operator selection on the CEC benchmark suite.
RL-CORCO~\cite{RL-CORCO} employs Q-learning to dynamically select two mutation operators for DE in constrained optimization problems. The population is divided into nine subpopulations based on the objective value and the degree of constraint violation, resulting in nine distinct states. Whenever a mutation strategy is applied and it either improves or maintains the performance of the population individuals, a reward of 1 is returned; otherwise, a reward of 0 is given. Additionally, RL-CORCO incorporates a population reinitialization mechanism to prevent it from becoming trapped in local optima. RL-CORCO demonstrates its superiority over other baseline methods on the CEC benchmark.

RL-HDE~\cite{RL-HDE} employs Q-learning to dynamically select six mutation operators and adjust two trigger parameters for DE. To select mutation operators, RL-HDE divides the population into 20 states based on diversity and average performance relative to the initial population. A reward 1 is given if a better solution is obtained with the selected mutation operator, 0 if there is no change, and -5 if the performance worsens. To balance global exploration and local exploitation, RL-HDE dynamically adjusts two hyperparameters and constructs six states based on a similar definition for operator selection.
Regarding rewards, a better solution yields a reward of 1, no change results in a reward of 0, and deterioration leads to a reward of -1. Experimental results demonstrate that RL-HDE outperforms other baselines in solving complex interplanetary trajectory design problems such as Cassini2 and Messenger-full.
DE-RLFR~\cite{DBLP:journals/swevo/LiSYSQ19} employs Q-learning to dynamically select one of three mutation operators for each individual in MOOP. Specifically, DE-RLFR categorizes the fitness of each objective into three levels based on their ranking, resulting in nine states for RL. A reward of 10 is assigned when offspring outperform their parents; otherwise, 0 is returned. Experiments across eleven multimodal multi-objective optimization problems demonstrate that DE-RLFR can effectively construct the superior Pareto front.
LRMODE~\cite{huang2020afitness} integrates the findings from a local landscape topology analysis with RL to approximate the optimal probability distribution for dynamically selecting MODE's mutation operators. 
LRMODE demonstrates superiority over other multi-objective optimization algorithms on multi-objective optimization tasks.
MOEA/D-DQN~\cite{tian2022deep} utilizes parent solutions and weight vectors as RL states and constructs RL rewards based on fitness improvements. It employs DQN to choose variation operators for MOEAs, leading to superior performance compared to other MOEAs across a diverse range of MOP benchmarks.
% MOEA/D-DQN employs the parent solutions and weight vectors as RL states, constructs RL rewards based on fitness improvements, and employs DQN to select variation operators for MOEAs. MOEA/D-DQN outperforms other MOEAs on a wide range of MOP benchmarks.
% MOEA/D-DQN将父代solution以及weight vectors作为RL state， 通过fitness improvements构建RL的奖励信号，使用DQN为MOEAs选择variation operators。MOEA/D-DQN在大量的MOP benchmark上优于其他MOEA算法。 
% MOEA/D-DQN utilizes DQN to choose crossover and mutation operators for the population in MOEAs, addressing multi-objective optimization problems.
AMODE-DRL~\cite{li2023scheduling} dynamically selects five mutation operators and adjusts two continuous parameters in multi-objective scheduling problems (MOSP). It leverages DQN to select mutation operators and DDPG to fine-tune continuous parameters. The RL states involve the current individual's fitness, fitness improvement, and population diversity. The RL rewards are defined by individual fitness and population diversity. Experimental results in both randomly generated instances and real-world problem cases demonstrate that DRL significantly enhances MODE's exploration and learning efficiency.
% AMODE-DRL动态调整五个mutation operators和两个连续的参数在多目标调度问题中。AMODE-DRL使用DQN来选择mutation operators，使用DDPG来调整连续参数。RL的状态被定义为：当前个体的fitness，适应度提升以及种群多样性，使用 the individual fitness
% and population diversity构建RL奖励信号。实验在 randomly generated instances and the practical problem instances表明DRL可以使得MODE更加高效地探索与学习。

% AMODE-DRL employs DDQN and DDPG to select mutation operators and parameter settings for MODE, effectively handling multi-objective scheduling problems.

%\textbf{基于RL的动态算法配置}

%EA算法在以往的研究中暴露出了两类问题：一，没有一类EA算子具备高效解决所有问题的能力，需要根据问题的特性以及专家知识进行EA算法选择。二，EA算法对超参数敏感，当不具备EA专业的调参经验时，超参数调整将是一个十分具有挑战的任务，微小的超参数变化会造成巨大的性能差异。
%为此一些工作尝试优化上述EA涉及到的算子选择以及超参数配置问题来提升EA的可用性与鲁棒性，这些工作被称为动态参数配置。在这里我们主要关注如何使用RL来辅助动态算法配置过程，我们将这些工作分为两个大的类别：动态算子选择 以及动态超参数配置。

% 动态算子选择：这类以EA为主的算法通常解决组合优化问题以及多目标优化问题。
%RL-GA将GA 与Q-learning结合起来通过Q值为每个个体选择交叉与变异算子用于提升GA解决旅行商问题的能力。
%RL-GA使用相同的技术用于解决electromagnetic detection satellite scheduling problem。
%不同于以往工作为每个个体选择交叉变异算子，
%EA+RL(O)动态的使用DDPG为整个种群选择交叉和变异算子，并在H-IFF优化问题以及旅行商问题上验证了方法的有效性。
%MARLwCMA则使用Q-learning选择DE中的变异算子，MPSORL使用Q-learning每一个particle选择变异算子。DE-DDQN则使用DDQN引入到了DE中，并未每一个父代选择变异算子。上述三个算法都在CEC上证明了RL可以通过选择算子提升不同EA算法的性能。
%RL-CORCO则使用Q-learning来为DE维护的种群动态地选择变异算子，并在车辆路由问题上证明了方法的有效性。除此之外，GSF将使用DQN和PPO用于intelligently selecting
%appropriate combinations of the algorithmic components (i.e. evolution operators) during different stages of the optimisation process. 并在车辆路由问题中证明了GSF的有效性与通用性。
%RL-HDE则使用Q-learning来为DE选择变异算子以及trigger parameters用于解决全局轨迹优化问题。MOEA/D-DQN使用DQN为MOEAs维护的种群选择交叉和变异算子来解决多目标优化问题。AMODE-DRL则使用DDQN和DDPG为MODE选择变异算子与参数设置来解决多目标调度问题。

\textbf{Dynamic Hyperparameter Configuration.}
In this context, we introduce the RL-assisted EA hyperparameter configuration, including crossover probability, mutation rate, population size, and others. AGA~\cite{DBLP:conf/esoa/EibenHKS06} leverages Q-learning to dynamically adjust the EA's crossover rate, mutation rate, tournament size, and population size. The RL states correspond to population information, e.g., maximum fitness, mean fitness, and the previous action vector.
The reward function is defined as the improvement of the best fitness.% incentivizing the RL algorithm to adjust parameters in a manner that leads to superior solutionsf
The experiments demonstrate that AGA outperforms GA on the Multimodal Problem Generator introduced by Spears~\cite{spears2000evolutionary}.
LTO~\cite{LTO} utilizes GPS~\cite{GPS} for dynamically adjusting the mutation step-size parameter of CMA-ES.
The RL states include the current step-size value, the current cumulative path length, the history of objective value changes, and the step-size history from the previous $h$ iterations.
LTO constructs the RL rewards based on the objective value of the current solution.
LTO demonstrates its efficiency in the BBOB-2009 benchmark~\cite{hansen2009real}.
%RL-DAC utilizes Q-learning to configure parameters across all algorithms. 
RL-DAC~\cite{DBLP:conf/ecai/BiedenkappBEHL20} formalizes DAC as a contextual MDP to enable RL to learn across a set of instances. It also introduces white-box benchmarks to demonstrate the efficiency of RL in hyperparameter tuning. Strictly speaking, RL-DAC is not limited to EAs; it can apply to all optimization algorithms.
REM~\cite{zhang2022variational} employs Variational Policy Gradient to continuously adapt the DE's scaling parameter and crossover rate. REM uses the present population information and the corresponding randomness as the state.
The reward gives 0 if the algorithm reaches the maximum generation. Alternatively, it provides the negative logarithm of the smallest function value discovered by the EA.
Experiments demonstrate that DE and adaptive DE, with tuned hyperparameters, outperform counterparts and other methods.
%like ParamILS, F-Race, and Bayesian optimization.

In hybrid EAs, the timing of switching between different EA phases is crucial for algorithm performance. Different from rule-based switching methods, Q-LSHADE and DQ-HSES~\cite{DBLP:journals/cim/ZhangSBZX23} propose an adaptive framework based on RL to adjust the switching time. Specifically, Q-LSHADE combines Q-learning and LSHADE~\cite{DBLP:conf/cec/TanabeF14}, adaptively controlling when to use the linear population size reduction (LPSR) technique within L-SHADE. DQ-HSES combines DQN and HSES~\cite{DBLP:conf/cec/ZhangS18a}, adaptively controlling when to transition from the univariate sampling phase to the CMA-ES algorithm.
%In the framework, the states are defined by the differences between the best individual and the neighboring generations, as well as the descent rate compared to the initial population. The action space $\mathcal{A}$ is $\{0, 1\}$, where $1$ means the use of LPSR or switching from the univariate sampling phase to CMA-ES, and $0$ indicates the opposite. If it's the terminal state (i.e., the agent reaches the horizon limit $T$ or the action is 1), the reward will be the negative logarithm of the minimum function value found so far. This encourages the RL agent to minimize the function value.
The experiments in the CEC 2014 and 2018 benchmarks demonstrate that the proposed algorithms outperform SOTA EAs.
% Q-LSHADE and DQ-HSES~\cite{DBLP:journals/cim/ZhangSBZX23} leverage Q-learning to determine whether to use linear population size reduction (LPSR) or not. The state encompasses two key components: the variation in the best function values across consecutive 50 generations and the rate of descent from the initial population.
% If it's the maximum generation or the agent chooses to implement LPSR, the reward will be the negative logarithm of the minimum function value found so far. Otherwise, the reward is zero.
% The experimental findings on the CEC 2014 and 2018 test suites demonstrate that the proposed algorithms exhibit substantial superiority over their counterparts, even surpassing some state-of-the-art EAs.
MADAC~\cite{DBLP:conf/nips/0001XYL0Z022}  emphasizes the heterogeneity among various hyperparameters and recognizes that applying a single RL algorithm for configuring all parameters can introduce complexities. Hence, MADAC applies a typical MARL method value-decomposition networks (VDN)~\cite{DBLP:conf/atal/SunehagLGCZJLSL18} to search for the appropriate settings for the multi-objective evolutionary algorithm MOEA/D's~\cite{zhang2007moea} four categories of hyperparameters. The RL states incorporate characteristics from different aspects, e.g., the specific problem instance, attributes associated with the ongoing optimization process, and aspects concerning the evolving population.
MADAC provides rewards when the algorithm discovers better solutions than the best so far and offers greater rewards to agents that can find even better solutions in later stages.
In multi-objective optimization challenges, MADAC demonstrates superior performance compared to other methods. qlDE~\cite{qlDE} uses Q-learning to dynamically adjust two hyperparameters of DE, the scale factor $F$ and crossover rate $Cr$. If the best individual in the population is better than the previous generation, the reward is 1; otherwise, it is -1.
qlDE demonstrates comparable or superior performance to other DE algorithms in five truss structural weight minimization problems. RLDE~\cite{hu2021reinforcement} employs Q-learning to dynamically adjust the scale factor $F$ of DE. Actions are defined as 0.0, -0.1, and 0.1, which are added to the current $F$. If the offspring is superior to the parent, the reward is 1; otherwise, it is 0. RLDE outperforms other algorithms in the parameter extraction problems involving various PV models.

\textbf{Others.} 
Here we introduce several methods within RL-assisted Optimization of EA in other aspects. 
{Some algorithms are influenced by the Baldwin effect and Lamarckian ideas~\cite{baldwin2018new, morgan1896modification}, which introduce learning into EAs to enhance evolutionary efficiency. The early work experimentally validates that the introduction of learning can improve the efficiency of EAs~\cite{hinton1987learning}. Subsequently, many works attempt to incorporate RL into EAs.
RGP~\cite{RGP} integrates Q-learning into tree-based GP to enhance evolutionary search. RGP utilizes GP to search trees, dividing the search space into coarse-grained regions. Q-learning is embedded at the leaf nodes of the tree for decision-making. Ultimately, RGP demonstrates that incorporating Q-learning can further improve the efficiency of GP in Maze tasks.
GNP-RL~\cite{GNP-RL} combines GNP~\cite{hirasawa2001comparison} with Q-learning. GNP leverages the higher expressive power and more compact graph structure to address the bloat issue of tree structure. GNP-RL employs RL to more fully exploit state and reward information returned by the environment, thereby enhancing optimization efficiency. Ultimately, GNP-RL demonstrates the efficiency of the method in grid environments.}

In addition, some works utilize RL to enhance EA-based planning methods.
%Model Predictive Control (MPC) is a well-known control algorithm. The core idea of MPC is to construct a world model and iteratively improve random action sequences through CEM to optimize planning, continuously refining the model. 
Model Predictive Control (MPC)~\cite{MPC} is a model-based control approach that begins by designing or learning a world model. Subsequently, it employs this model to plan a sequence of actions. To enhance efficiency, several works replace traditional random sampling planning methods with CEM, such as PETS~\cite{Pets}, PlaNet~\cite{PlaNet}, and POPLIN~\cite{Poplin}. 
Building upon CEM, some works incorporate RL or gradient optimization into MPC to enhance performance.
In Grad-CEM~\cite{Grad-CEM}, several random action sequences are generated, and stochastic gradients obtained from maximizing rewards based on the dynamic model are used to update the generated sequences, which improves the efficiency of CEM. Experiments in MuJoCo and DMC benchmarks  demonstrate that Grad-CEM outperforms CEM.
LOOP~\cite{LOOP} combines MPC and off-policy RL. To enhance estimation accuracy,  LOOP augments the traditional H-step discounted rewards with Q-values. Additionally, trajectories generated by the RL policy based on the world model are combined with those generated by CEM to optimize the CEM distribution. LOOP outperforms other model-based methods on MuJoCo tasks.
TD-MPC~\cite{TD-MPC} employs the same framework as LOOP, with the distinction of encoding states into a latent space for modeling the world model, learning the policy, and approximating Q-values.
Experiments show that TD-MPC outperforms LOOP and SAC on DMC tasks.

% 这里我们介绍几个EA与RL在其他方面的结合。MPC是一个著名的控制算法，MPC的核心思想是根据构建的dynamic模型，通过CEM对随机采样的动作序列进行迭代提升，并不断优化模型来更好的进行规划。很多工作都使用了MPC并在控制问题中得到应用，例如PETS，PlaNet，POPLIN。
% 一些工作也尝试将RL或者梯度优化融入到MPC中提升性能。Grad-CEM随机生成N个动作序列, 并基于dynamic model最大化reward获得的随机梯度来更新随机生成的动作序列，并选取top k来更新CEM，提升CEM的优化效率。
% TD-MPC则维护了一个额外的RL训练的Critic和Actor，一方面将H步的累计折扣奖励替换为 H步的累计折扣奖励加上Q值来提升准确度，另一方面将由RL优化的策略生成的轨迹加入到CEM生成的轨迹中优化CEM。TD-MPC在DMC上显示出了明显的性能增益。

%除此之外，CEM也被应用在了控制领域，例如PETS，PlaNet，POPLIN，TD-MPC都使用了MPC，而MPC是通过构建的world model来进行短期的轨迹收集，通过优化短期轨迹来近似全局最优策略。而很多时候采用CEM来进行控制动作的选择。
% TD-MPC则将RL引入到了MPC的学习过程中，使用Q value提到未来估计，同时使用RL学到的策略来进行轨迹生成，与CEM生成的轨迹一起用于top K轨迹的选取，优化策略与模型。

\textbf{Challenges and Future Directions:}
The above works demonstrate the efficiency of RL-assisted EAs in various aspects. 
%Compared to the EA-assisted RL direction, the direction of RL-assisted EAs not only focuses on SDP but also encompasses a wide range of optimization problems, such as CTOP and COP.
Despite demonstrating the capability of RL to enhance EAs across various types of problems, RL-assisted EAs still face the following challenges:
1) Utilizing RL-assisted Optimization of EA requires researchers to have a deep understanding of the target problem to formulate it as an MDP. 
Additionally, RL knowledge is necessary to select suitable algorithms for learning.
%This demands a rich knowledge of RL and a thorough comprehension of the problem.
2) RL introduces extra hyperparameters, which usually need to be adjusted based on the specific problem. This may entail additional trial-and-error overhead.
3) Although RL has demonstrated the ability to enhance EAs in experiments across different branches, this lacks theoretical support and convergence guarantees.
4) Despite employing similar techniques, the absence of comparisons between different methods, especially in dynamic algorithm configuration, makes it challenging to determine which method currently outperforms in addressing specific problems.
Based on the aforementioned challenges, we propose several future research directions:
1) More advanced and stable RL algorithms within the EA process require further investigation, e.g., exploring more generalized modeling approaches and developing more robust and general RL algorithms.
2) Establish theoretical guarantees for RL-assisted EAs, including convergence and performance bounds.
3) For each research branch, further investigation can be conducted to address existing limitations of current methods, e.g., develop more efficient population evaluation methods and mutation operators.
4) Researchers can construct a unified framework and evaluate the related works in a consistent benchmark, offering more valuable insights.
% 2) Similar to EA, RL hyperparameter tuning demands careful settings. 
% 3) The role of more advanced and stable RL algorithms within the EA process requires further investigation.
% Therefore, future research directions can focus on addressing these challenges. For instance, exploring more generalized modeling approaches and developing more robust and versatile RL algorithms would be worthwhile avenues of investigation.

% 动态超参数配置：这里主要涉及的超参数包括：交叉的概率，变异率，种群大小等，AGA将参数选择建模为MDP过程，并使用Q-learning为EA dynamic control the crossover rate, mutation rate, tournament size and population size. LTO则使用GPS来为CMA-ES dynamically configure the mutation step-size parameter. RL-DAC则使用Q-learning为所有算法配置
% RL-DAC 形式化DAC as a contextual MDP to make RL can learn across a set of instances,并提出了一个白盒优化基准证明了RL在超参数调整上的有效性。值得注意的是RL-DAC不只局限于EA算法，而是所有优化算法。
%RL-DAC使用Q-learning和DDQNxxxxxx.
%REM使用 policy gradient为DE来动态调整变异率和交叉的概率。Q-LSHADE和DQ-HSES分别将Q-learning和Deep Q-learning与 LSHADE和HSES结合来控制混合EA算法中的从一个EA算法跳转到另一个EA算法的switching time。MADAC则认为不同的参数之间存在异构的问题，简单的使用单个RL算法配置所有参数会引入困难。为此，MADAC使用多智能体算法VDN来为MOEA/D动态配置dynamic control  weights, neighborhood size,and reproduction operator type, Parameters of reproduction operators参数。MADAC同时涉及到了动态超参数配置与算子选择。

\begin{table*}[t]
  \renewcommand{\arraystretch}{1.4}
  \centering
  \caption{Synergistic Policy Optimization with EA and RL in Single-Agent Settings.}
  \begin{tabular}{lllllllllll}
    \toprule
    \multicolumn{8}{c}{\textbf{Both EAs and RL search in policy space}} \\
    \midrule
    \textbf{Algorithm} &  \textbf{Task} &\textbf{EA} & \textbf{RL}  & \makecell[l]{\textbf{Fitness}\\ \textbf{Surrogate}}  & \makecell[l]{\textbf{Policy} \\ \textbf{Structure}} & \textbf{RL Role} & \textbf{EA Role}  \\
    \midrule
    \multirow{1}{*}{ERL~\cite{khadka2018evolution}} &  MuJoCo & GA & DDPG   & N/A & Private & \makecell[l]{Policy Injection} & \makecell[l]{Diverse Experiences For RL}  \\
    % \midrule
    % \multirow{1}{*}{X-DDPG~\cite{DBLP:conf/mod/EspositiB20a}} & Gym Tasks & GA & DDPG   &  & Private   & \makecell[l]{Policy\\ Injection} & \makecell[l]{Diverse Experiences \\ For RL Optimization}   \\
        \midrule
    \multirow{1}{*}{CERL~\cite{khadka2019collaborative}} &   MuJoCo & GA & TD3 &  N/A & Private  & \makecell[l]{Policy Injection} & \makecell[l]{Diverse Experiences For RL}    \\
    \midrule
    \multirow{1}{*}{PDERL~\cite{bodnar2020proximal}} &   MuJoCo & PD-GA & DDPG &  N/A & Private  & \makecell[l]{Policy Injection} & \makecell[l]{Diverse Experiences For RL}    \\
     \midrule
    % \multirow{1}{*}{CEM-RL~\cite{pourchot2018cem}} &   MuJoCo & CEM & TD3 & & Private  & \makecell[l]{Gradient Injection} & \makecell[l]{}    \\
    %  \midrule
    % \multirow{1}{*}{Supe-RL~\cite{marchesini2021genetic}} &   MuJoCo & GA & PPO \& Rainbow & & Private  & \makecell[l]{Provide Elite} & \makecell[l]{Learning Guidance}    \\
    %  \midrule
    \multirow{1}{*}{SC~\cite{wang2022surrogate}} &  MuJoCo & \makecell[l]{GA \&\\ PD-GA} & DDPG  & \makecell[l]{Using Critic \\Estimates} & Private  & \makecell[l]{Policy Injection} & \makecell[l]{Diverse Experiences For RL}  \\
     \midrule
    \multirow{1}{*}{GEATL~\cite{GEATL}} & \makecell[l]{Grid World} & GA & \makecell[l]{A2C}  &   N/A & Private   & \makecell[l]{Policy Injection} & \makecell[l]{Elite Policy Synchronisation}\\
    \midrule
    \multirow{1}{*}{CSPS~\cite{DBLP:conf/nips/ZhengW0L0Z20}} &   MuJoCo &CEM & \makecell[l]{PPO \& \\SAC}  &  N/A & Private   & \makecell[l]{Policy Injection} & \makecell[l]{Diverse Experiences For RL \\Two Separated Buffer}  \\
    \midrule
    \multirow{1}{*}{T-ERL~\cite{DBLP:conf/gecco/ZhengC23}} & \makecell[l]{ MuJoCo} & ES & \makecell[l]{TD3}  &  N/A & Private   & \makecell[l]{Policy Injection} & \makecell[l]{Diverse Experiences For RL\\ Two Separated Buffers}\\
    \midrule
    \multirow{1}{*}{ESAC~\cite{DBLP:conf/atal/Suri22}} &   \makecell[l]{ MuJoCo \\ \& DMC } & A-ES & SAC  &  N/A  & Private  & \makecell[l]{Policy Crossover} & \makecell[l]{Diverse Experiences For RL}   \\
    \midrule
    \multirow{1}{*}{PGPS~\cite{kim2020pgps}} &   \makecell[l]{ MuJoCo } & CEM & TD3 & \makecell[l]{Using Critic \\Estimates} & Private   & \makecell[l]{Gradient Injection} & \makecell[l]{Diverse Experiences For RL \\ \& Guided Policy Learning}  \\
    % \multirow{1}{*}{PBRL~\cite{pretorius2021population}} & \makecell[l]{ MuJoCo} & GA & \makecell[l]{DDPG}  &  & Private   & \makecell[l]{Policy\\ Injection} & \makecell[l]{Diverse Experiences \\ For RL Optimization\\ Two Separated Buffers}\\
    % \midrule
    \midrule
    \multirow{1}{*}{ERL-Re$^2$~\cite{Re2}} & \makecell[l]{ MuJoCo \\ \& DMC } & B-GA & \makecell[l]{DDPG \\\& TD3}  & \makecell[l]{H-Step \\Bootstrap (PeVFA)} & Shared   & \makecell[l]{Policy Injection} & \makecell[l]{Diverse Experiences For RL} \\
    \midrule
    \multirow{1}{*}{VEB-RL~\cite{li2023value}} & \makecell[l]{ MinAtar \\ \& Atari } & \makecell[l]{GA \&\\CEM} & \makecell[l]{DQN \&\\ Rainbow}  & \makecell[l]{The TD Error} & Private   & \makecell[l]{Value Function \\ Injection} & \makecell[l]{Diverse Experiences For RL} \\
    \midrule
    \multirow{1}{*}{EvoRainbow-Exp~\cite{lievorainbow}} & \makecell[l]{ MuJoCo \\ \& Maze \& \\ MinAtar \& \\ MetaWorld } & CEM & \makecell[l]{TD3 \\\& SAC}  &  N/A & Private   & \makecell[l]{Policy Injection} & \makecell[l]{Diverse Experiences For RL \\ Genetic Soft Update} \\
    \midrule
    \multirow{1}{*}{EvoRainbow~\cite{lievorainbow}} & \makecell[l]{ MuJoCo \\ \& Maze \& \\ MinAtar \& \\ MetaWorld } & CEM & \makecell[l]{TD3 \\\& SAC}  & \makecell[l]{H-Step \\Bootstrap (Critic)} & Shared   & \makecell[l]{Policy Injection} & \makecell[l]{Diverse Experiences For RL \\ Genetic Soft Update} \\
    \midrule
    \multicolumn{8}{c}{\textbf{EAs search in solution space \& RL search in policy space}} \\
    \midrule
    \textbf{Algorithm} & \textbf{Task} & \textbf{EA} & \textbf{RL} & \multicolumn{2}{c}{\textbf{Solution Representation}} & \textbf{RL Role} & \textbf{EA Role} \\
    \midrule
    CORE~\cite{li2025core} & Floorplanning & CDES & PPO & \multicolumn{2}{c}{B$^\ast$\mbox{-}Tree} & \makecell[l]{Solution Injection} & \makecell[l]{Corrects RL\\optimization} \\
    \bottomrule
  \end{tabular}
\end{table*}

\section{Synergistic Optimization of EA and RL}
\label{sec: parallel}

% \begin{figure}[t]
% \centering
% % \subfigure[Sequential Integration of EA and RL.]{
% % \centering
% % \includegraphics[width=0.8\linewidth]{Survey/EA_RL_1.pdf}
% % \label{Sequential Integration of EA and RL for Optimization}
% % }
% \includegraphics[width=0.8\linewidth]{Survey/EA_RL_2.pdf}
% %\vspace{-0.15cm}
% \caption{Schematic for Synergistic Optimization of EA and RL.}
% %\label{ablation study}
% \label{Schematic for Synergistic Optimization of EA and RL.}
% \end{figure}

The previous hybrid algorithms typically maintain only one of the approaches (EAs and RL) as the primary problem solver, while the other algorithm plays a supporting role. 
{This section focuses on synergistic optimization algorithms that integrate the complete learning and optimization processes of RL and EAs, either (1) to simultaneously solve the same problem with collaborative mechanisms, or (2) independently optimize subproblems to obtain partial solutions, which are subsequently combined to form a complete solution.
The schematics are illustrated in
Figure~\ref{Schematic for Synergistic Optimization of EA and RL A.} and Figure~\ref{Schematic for Synergistic Optimization of EA and RL B.}.}
The related works in this direction focus on SDP. 
Below, we separately introduce two different collaboration approaches.
%In these problems, EAs and RL need to learn a control policy collaboratively to obtain higher returns.
%Below, we will separately introduce two different collaboration approaches along with their related works.}

{The first collaboration approach involves simultaneously solving the same problem using EAs and RL, with collaboration during the solving process. 
This collaboration approach is inspired by the complementary strengths demonstrated by EAs and RL.
Specifically, 
EAs, based on population and random exploration, offer excellent exploration and global optimization capabilities~\cite{khadka2018evolution}.
However, random search in vast parameter space often leads to low optimization efficiency. Additionally, EAs evaluate individuals based on episodic rewards, which necessitate each individual to interact with the environment for fitness. These weaknesses result in significant sample costs, often ranging two to three orders of magnitude higher compared to RL~\cite{salimans2017evolution}. While these coarse-grained episodic rewards make EAs more insensitive to the quality of the reward signals. In contrast, RL can leverage finer-grained information, e.g., states and rewards, and reuse historical experiences, thereby providing higher sample efficiency, yet it suffers from exploration challenges during the learning process, often prone to converging to suboptimal solutions. Moreover, RL necessitates well-designed reward signals to ensure final performance~\cite{survey2}. 
Through the comparison above, we observe that EAs and RL each have their strengths and weaknesses. The key point of this collaboration approach is how to establish a symbiotic relationship, maintaining their respective strengths while compensating for their weaknesses.} Consequently, many works try to integrate EAs with RL for synergistic optimization to enhance search efficiency and solution quality.

\textbf{Single-Agent Optimization.} The earliest method is ERL~\cite{khadka2018evolution}, which establishes the foundational framework.% for synergistic optimization of EA and RL. 
%As depicted in Figure X, 
In ERL, both EAs and RL engage in policy search. EAs provide the diverse samples generated during population evaluation to RL for policy learning, thereby enhancing sample efficiency. Conversely, RL incorporates its policy into the population to participate in the evolutionary process. If the RL policy achieves better performance, it guides and facilitates population evolution. Through these mechanisms, ERL integrates the strengths of both EAs and RL. Experimental results demonstrate that ERL outperforms DDPG and GA on most OpenAI MuJoCo tasks.
CERL~\cite{khadka2019collaborative} is a follow-up work to ERL, focusing on solving the sensitivity problem to RL discount factors $\gamma$. It is important to note that the role of GA in CERL is not employed for hyperparameter tuning but for policy search, which is consistent with that in ERL. Thus we discuss it in this section.
CERL maintains multiple RL learners with different gammas. Unlike dynamic adjustments, CERL predefines the gamma values without tuning them in the learning process. During training, resources are dynamically allocated based on the performances of learners. Experiments on MuJoCo tasks demonstrate that CERL is more insensitive to hyperparameters.
Taking inspiration from GPO~\cite{GangwaniP18GPO}, PDERL~\cite{bodnar2020proximal} proposes Proximal-Distilled GA (PD-GA) to address the catastrophic forgetting issue associated with GA in ERL.
Specifically, PD-GA encompasses novel crossover and mutation operators. 
The crossover operator distills desirable behaviors from parents to offspring based on the Q values. The mutation operator adjusts the magnitude of mutations by taking into account parameter sensitivity to actions. PDERL demonstrates superior performance to ERL in OpenAI MuJoCo tasks.
SC~\cite{wang2022surrogate} focuses on mitigating the high sample cost associated with population evaluation. It proposes leveraging RL critic as a surrogate for fitness and evaluating individuals with the critic based on the samples from the replay buffer. Besides, SC introduces two mechanisms for surrogate utilization: 1) employing the surrogate for population evaluation with a probability of $P$, while interacting with the environment with a probability of $1-P$; 2) generating a population larger than twice the original size and then using the surrogate model to filter half of the individuals. SC integrates with ERL and PDERL, demonstrating performance improvements on MuJoCo tasks.

Unlike previous approaches that integrate off-policy RL with EA, GEATL~\cite{GEATL} combines on-policy RL with EA. Similar to ERL, in GEATL, RL influences EAs through policy injection. However, the influence of EAs on RL operates differently: when the elite policy of the population outperforms the RL policy, the elite policy replaces the RL policy. Moreover, if their performances are comparable, there is a 50\% chance that the elite policy replaces the RL policy. GEATL demonstrates its superiority over ERL in scenarios in Grid World with sparse rewards.
%Previous frameworks typically focus on integrating either on-policy RL or off-policy RL with EA. 
CSPC~\cite{DBLP:conf/nips/ZhengW0L0Z20} incorporates off-policy RL, on-policy RL, and EAs. %, aiming to fuse the strengths of three algorithms with distinct update principles.
Specifically, CSPS integrates SAC, PPO, and CEM. When the SAC policy outperforms PPO or the policies in the population, it replaces those individuals. Similarly, if the PPO policy excels, it replaces the population policies. Furthermore, CSPS introduces an additional local experience buffer for SAC to store recently generated experiences and incorporates several experience filtering mechanisms. These mechanisms ensure that the added local experiences are the most recent and superior to the minimum value among all individuals at that time. SAC utilizes the local experience buffer for policy optimization with a probability of $P$, while utilizing the global buffer with a probability of $1-P$. %By capitalizing more on recent experiences, SAC can enhance optimization efficiency. 
CSPS outperforms three basic algorithms on most of the MuJoCo tasks.

T-ERL~\cite{DBLP:conf/gecco/ZhengC23} integrates ES with TD3 and constructs two replay buffers akin to CSPS. One buffer saves the experiences of all individuals, while the other saves recent RL experiences. T-ERL proportionally samples from both buffers for RL training. T-ERL demonstrates superiority over TD3 on three of four MuJoCo tasks
ESAC~\cite{DBLP:conf/atal/Suri22} adopts the ERL framework while replacing DDPG with SAC and GA with a modified ES. The ES introduces an automatic adjustment mechanism to regulate the coefficient of added Gaussian noise in ES, denoted as A-ES. The coefficient is updated based on the disparity between the best performance identified within the population and the average performance. Moreover, unlike GA, ESAC does not shield elite individuals from mutation interference; instead, it employs crossover between elites and the updated ES distribution to transmit favorable traits from elites to the offspring. ESAC demonstrates its superiority over ES and other RL algorithms on MuJoCo and DMC tasks.
PGPS~\cite{kim2020pgps} follows the ERL framework to combine CEM and TD3. It's noteworthy that PGPS maintains a full life-cycle RL policy, distinguishing itself from CEM-RL. Within PGPS, the population of size $N$ consists of the elite from the previous generation (index 0), individuals randomly sampled from the CEM distribution (index 1 to $\frac{N}{2}$), and individuals selected from the large CEM-sampled pool using the surrogate mechanism from SC (index $\frac{N}{2}$ to $N$). Moreover, PGPS introduces Guided Policy Learning. When the behavior difference between the elites in the population and the current RL actor exceeds a threshold, behavior cloning is employed to assist RL learning. Conversely, the constraints are relaxed if the difference is within the threshold. PGPS demonstrates superior or comparable performance to CEMRL, PDERL, CERL, CEM, and some RL algorithms in MuJoCo. 

ERL-Re$^2$~\cite{Re2} uncovers a primary problem prevalent in the existing ERL-related research: the wide use of isolated policy architectures, where each individual operates within its private policy network. However, this independent structure often hinders the efficient transfer of valuable knowledge. To solve the problem, ERL-Re$^2$ decomposes the policies into a shared state representation and independent linear policy representations. The policy structure facilitates knowledge sharing while simultaneously compacting the policy space. Moreover, ERL-Re$^2$ proposes behavioral-level genetic operators (B-GA) based on linear policy representations, coupled with an $H$-step bootstrap fitness surrogate for population evaluation. ERL-Re$^2$ achieves state-of-the-art performance on MuJoCo tasks.

% VEB-RL~\cite{li2023value}主要用于解决以往工作忽视value-based RL的问题，VEB-RL构建了一个种群的value function和相应的target function，并使用TD error作为Fitness来进行值函数近似。
% VEB-RL还引入了精英交互机制避免交互资源浪费。VEB-RL在MinAtar和Atari上显著提升了DQN和Rainbow。

VEB-RL~\cite{li2023value} addresses the issue of previous works overlooking value-based RL. VEB-RL constructs a population of value functions and corresponding target functions, using negative TD error as the fitness metric for value function evaluation. VEB-RL also introduces an elite interaction mechanism to avoid wasting interaction resources. VEB-RL significantly enhances DQN and Rainbow on MinAtar and Atari.

% EvoRainbow和EvoRainbow-Exp系统性的通过五个角度实验回顾了这一分支的工作，详细对比了相同功能的机制，并通过最有效的机制构建出了EvoRainbow和其探索版本EvoRainbow-Exp，EvoRainbow融合了平行mode，共享架构，CEM，Genetic Soft Update以及H-Step Bootstrap(Critic)。EvoRainbow则融合了平行mode，私有架构，CEM和Genetic Soft Update。 EvoRainbow和EvoRainbow-Exp在locomotion tasks，maniplation tasks，Maze tasks，Minatar等20个任务上证明了优于目前ERL的SOTA方法。

EvoRainbow~\cite{lievorainbow} and EvoRainbow-Exp~\cite{lievorainbow} systematically review this branch of works from five perspectives through experiments, providing a detailed comparison of mechanisms with the same functionality. By integrating the most effective mechanisms, they construct EvoRainbow and its exploratory version, EvoRainbow-Exp. EvoRainbow incorporates parallel mode, shared architecture, CEM, Genetic Soft Update, and H-Step Bootstrap (Critic). EvoRainbow-Exp combines parallel mode, private architecture, CEM, and Genetic Soft Update. Both EvoRainbow and EvoRainbow-Exp have demonstrated superiority over the current state-of-the-art ERL methods across 20 tasks, including locomotion tasks, manipulation tasks, Maze tasks, and Minatar.

All of the aforementioned works focus on policy search, where both RL and EAs operate in the policy parameter space to optimize policies. However, in many combinatorial optimization problems, the ultimate goal is to obtain a solution rather than a policy. In such cases, EAs are particularly suitable, as they can directly explore the solution space without relying on search in a high-dimensional policy space.
To address constrained combinatorial optimization problems, CORE~\cite{li2025core} proposes a novel EA–RL hybrid framework. In CORE, EAs conduct clustering-based evolutionary search directly in the solution space, while RL (implemented with PPO) explores the policy parameter space.
To enable synergistic optimization, CORE transforms high-quality solutions found by EAs into decision sequences for RL, and uses these to construct constraint terms that steer the RL optimization direction.
Conversely, RL generates solutions based on the learned policy, which are then injected back into the EA population to enhance evolutionary progress.
Experimental results show that CORE outperforms both standalone EA and RL baselines on complex floorplanning tasks.

\begin{table*}[t]
  \renewcommand{\arraystretch}{1.4}
  \centering
  \caption{Synergistic Policy Optimization with EA and RL in Multi-Agent Settings.}
  \begin{tabular}{lllllllllll}
    \toprule
    \textbf{Algorithm} & \textbf{Task and Setting}  & \textbf{EA} & \textbf{MARL} & \textbf{EA Role} & \textbf{RL Role} & \textbf{Other features} \\
    % \midrule
    % \multirow{1}{*}{COMIX~\cite{COMIX}} & \makecell[l]{MA-MuJoCo\\ (Only Team rewards \\ Partial observation)} & CEM & QMIX &  Action Selection &  \makecell[l]{Value Function \\Approximation}   &  \makecell[l]{Composite\\ Structure} \\
 %    \midrule 
 %    \multirow{1}{*}{ROMANCE~\cite{ROMANCE}} &  \makecell[l]{SMAC\\ (Only Team rewards \\ Partial observation)} & QD & QMIX & \makecell[l]{Optimize 
 % and Maintain \\Diverse Attackers} &  \makecell[l]{Optimize Collaborators}  & \makecell[l]{Robust \\ MARL} \\ 
 %    \midrule 
 %    \multirow{1}{*}{MA3C~\cite{MA3C}} &  \makecell[l]{SMAC, Traffic Junction\\, Gold Panner, Hallway} & GA-like & CMARL & \makecell[l]{Optimize 
 % and Maintain \\Diverse Attackers} &  \makecell[l]{Optimize Collaborators}  & \makecell[l]{Robust \\ Communication} \\ 
 %    \midrule 
 %    \multirow{1}{*}{EPC~\cite{EPC}} &  \makecell[l]{ Particle-world \\Environment} & GA-like & MADDPG & \makecell[l]{Search The Team Policy\\ with Better Adaptation} &  \makecell[l]{Optimize MARL\\ with Different Scale}  & \makecell[l]{Large-Scale MALR} \\ 
    \midrule
    \multirow{1}{*}{MERL~\cite{MERL}} & \makecell[l]{MPE (Global Information \\ \& Dense agent reward \& \\sparse team reward }& GA & MATD3 & \makecell[l]{Optimize Population \\with Team Reward \& \\ Provide Experiences}   & \makecell[l]{Optimize MARL \\with Agent Reward \\ \& Inject MARL Policy}  & \makecell[l]{Two Types \\of Rewards} \\
    \midrule
    \multirow{1}{*}{NS-MERL~\cite{NS-MERL}} & \makecell[l]{Multi-rover Exploration\\ Domain (Global Information \\ \& Dense agent reward \& \\sparse team reward }& GA & MATD3 & \makecell[l]{Optimize Population \\with Team Reward \& \\ Provide Experiences}   & \makecell[l]{Optimize MARL \\with Agent Reward \\ \& Inject MARL Policy}  & \makecell[l]{Two Types \\of Rewards \\ \& Exploration} \\
        \midrule
    \multirow{1}{*}{CEMARL~\cite{CEMARL}} & \makecell[l]{MPE (Global Information \\ \& Dense agent reward \& \\sparse team reward }&  CEM & MATD3 & \makecell[l]{Optimize Population \\with Team Reward \& \\ Provide Experiences}   & \makecell[l]{Optimize MARL \\with Agent Reward}  & \makecell[l]{Two Types \\of Rewards} \\
     \midrule
    % \multirow{1}{*}{MAEDyS} & \makecell[l]{The Rover Domain \\ (Partial Information \\ \& Dense agent reward \& \\sparse team reward } & \\
    % \midrule
    \multirow{1}{*}{EMARL~\cite{EMARL}} &  \makecell[l]{The Flocking Env\\ (Only Team rewards \\ Partial observation)} & GA & COMA & \makecell[l]{Optimize Population \\with Team Reward \& \\ Provide Experiences} &  \makecell[l]{Optimize Population \\with Team Reward}  & \makecell[l]{Serial \\ Optimization} \\ 
 %     \midrule 
 %    \multirow{1}{*}{Mix-ME~\cite{Mix-ME}} &  \makecell[l]{MA-MuJoCo} & MAP-Elites & CMARL & \makecell[l]{Optimize 
 % and Maintain \\Diverse Attackers} &  \makecell[l]{Optimize Collaborators}  & \makecell[l]{Robust \\ Communication} \\ 
    \midrule 
    \multirow{1}{*}{RACE~\cite{li2023race}} & \makecell[l]{SMAC \& MA-MuJoCo\\ (Only Team rewards \\ Partial observation)} & A-GA & \makecell[l]{FACMAC\\or MATD3} &  \makecell[l]{Optimize Population \\ with Team Reward \& \\ Provide Experiences} & \makecell[l]{Optimize MARL \\ with Team Reward \\ \& Inject MARL Policy}    & \makecell[l]{Shared \\ Representation \\ Architecture}  \\
    \bottomrule
  \end{tabular}
\end{table*}

\textbf{Multi-Agent Optimization.}
In addition to the aforementioned single-agent optimization methods, the fusion of EAs and Multi-Agent Reinforcement Learning (MARL) has also made many advances. 
Here, the focus is on cooperative settings, where we need to control multiple agents to complete tasks.
Compared to MARL, EAs offer additional advantages: EAs avoid the need to model the MARL problem as the MDP, thus circumventing the non-stationarity problem~\cite{survey1,li2023race}.
% Among these, COMIX~\cite{COMIX} is an early and influential work in this area, which aims to address the problem of QMIX's inapplicability in continuous action spaces.
% Strictly speaking, COMIX falls under the category of EA-assisted RL action selection, rather than synergistic optimization. However, due to the multi-agent setup, we will introduce it in this section.
% % COMIX does not belong to the category of synergistic optimization, as it involves the simultaneous integration of both EA (CEM) and RL (QMIX) for decision-making and optimization.
% Within COMIX, QMIX is utilized for value function approximation, followed by CEM for action selection. Notably, COMIX outperforms MADDPG and IQL in tasks such as SMAC and MA-MuJoCo.
Among these, MERL~\cite{MERL} is proposed to efficiently utilize both team-level and agent-level rewards for collaboration. MERL maintains a team population and optimizes the team policies using EAs with team rewards, while simultaneously optimizing individual policies using RL with agent rewards. The overall optimization process is similar to ERL. MERL demonstrates superior performance to MATD3 and MADDPG on MPE tasks. 
NS-MERL~\cite{NS-MERL} extends MERL by considering two types of rewards during the optimization of individual rewards. To encourage exploration, NS-MERL employs a count-based method to track the number of times the current observation has been visited, where higher counts result in lower rewards. This reward is then multiplied by the original heuristic reward. Additionally, a counterfactual mechanism is utilized to calculate the contribution of each individual, thereby enhancing collaboration. The final reward is determined by multiplying the counterfactual reward with the two aforementioned rewards. Experimental results demonstrate that the constructed reward function outperforms other reward functions in the multi-rover exploration domain.
CEMARL~\cite{CEMARL} shares the same idea as MERL, primarily replacing GA with CEM. In each iteration, the population teams are sampled from the CEM distribution. Subsequently, a random individual is selected and optimized using MARL based on individual rewards. Following individual optimization, all teams interact with the environment to obtain team rewards, which are then used to optimize the CEM distribution. CEMARL also maintains a policy that is soft-updated to the team with the best performance in the population, enhancing stability. Experiments show that CEMARL outperforms MERL in MPE environments.
CEMARL does not maintain a full life-cycle MARL policy. Instead, it samples the MARL policy from the CEM distribution each time. Therefore, we can view MARL as the variation operator, injecting gradients into one individual of the population. As CEMARL aligns more closely with MARL settings, we include it in this branch for discussion.
Different from MERL and CEMARL, EMARL~\cite{EMARL} focuses on more general task settings, i.e., only team-level rewards. EMARL combines GA with COMA, where the population individuals are first optimized using GA, and the optimized population is further enhanced using policy gradients. EMARL is evaluated on the flocking tasks and shows better performance compared to benchmark MARL algorithms. 
The aforementioned algorithms are evaluated on simple tasks, whereas RACE~\cite{li2023race} proposes a new hybrid framework and demonstrates its efficiency in facilitating collaboration within complex tasks for the first time. Specifically. RACE introduces the concept of shared representations into the integration of MARL and EA. 
RACE divides the team policy into shared observation encoders and independent linear policy representations.
RACE maximizes the value function and Value-Aware Mutual Information to inject collaboration-related information and superior global states into the shared representations. The agent-level crossover and mutation operations are operated on the linear representations to ensure stable evolution. Finally, RACE achieves superior performance compared to FACMAC, MATD3, and MERL on SMAC and MA-MuJoCo tasks.

{In addition to the above collaboration approach, we can also decompose the problem into subproblems suited for EAs and subproblems suited for RL. A common pattern based on this collaboration approach is to utilize EAs for structure search and RL for policy learning. Next, we systematically introduce methods involving this collaboration approach.}

\begin{table*}[t]
  \renewcommand{\arraystretch}{1.4}
  \centering
  \caption{Synergistic Policy Optimization with EA and RL in Reward Design.}
  \begin{tabular}{lllll}
    \toprule
    \textbf{Algorithm} & \textbf{Task and Setting}  & \textbf{EA} & \textbf{MARL/RL} & \textbf{Other features} \\
    \midrule
    Evo-Reward~\cite{Evo-Reward} & Hungry--Thirsty task & PushGP & Q-learning & No LLM used \\
    \midrule
    Eureka~\cite{Eureka} & Manipulation \& Locomotion & LLM-based GA & PPO & LLM-driven reward design \\
    \midrule
    DrEureka~\cite{Dreureka} & Real-world Locomotion & LLM-based GA & PPO & Adds domain randomization; real-robot deployment \\
    \midrule
    Roska~\cite{ROSKA} & Manipulation \& Locomotion & LLM-based GA & PPO & Inter-generation policy inheritance \\
    \midrule
    R*~\cite{li2025r} & Manipulation & LLM-based GA & PPO & Reward-function crossover \& parameter optimization \\
    \midrule
\multirow{1}{*}{LaRes~\cite{li2025lares}} 
& \makecell[l]{Manipulation \& Locomotion\\ \& MinAtar} 
& LLM-based GA 
& SAC \& DQN 
& \makecell[l]{Experience sharing \& continual policy learning} \\
    \midrule
    LERO~\cite{wei2025lero} & MPE & LLM-based GA & MAPPO, VDN, and QMIX & Hybrid reward credits \& observation enhancement \\
    \midrule
    ReMAC~\cite{li2025remac} & Multi-Agent Manipulation & LLM-based GA & MASAC & Multi-agent reward disentanglement \\
    \bottomrule
  \end{tabular}
\end{table*}

\textbf{Reward Design}: 
Reward design is a crucial step in solving a task. Given a task, we typically first construct a reward function, and then select an appropriate algorithm to solve it based on this reward function. Therefore, the overall problem-solving process can often be decoupled into two components: reward design and policy learning. The former is usually addressed using EAs, while the latter is typically handled by RL.

Early works on reward design often relied heavily on manually predefined templates, within which the reward function was searched and optimized. A representative work is Evo-Reward~\cite{Evo-Reward}. Evo-Reward constructs a population of reward functions and employs PushGP~\cite{PushGP}, a variant of GP, for population evolution. Experimental results indicate that Evo-Reward can discover more efficient reward functions than the original reward function on the Hungry–Thirsty task.
With the rapid advancement of large language models (LLMs), recent studies have explored leveraging their coding capabilities and domain knowledge to design reward functions. Eureka~\cite{Eureka} generates a population of reward functions using LLMs. The task is then solved using PPO, and the final policy performance is adopted as the fitness of each reward function. Based on the reflection mechanism of LLMs, the reward population is iteratively improved. Eureka demonstrates performance that surpasses human-designed reward functions across more than ten manipulation and locomotion tasks.
DrEureka~\cite{Dreureka} extends this line of work by using LLMs not only to write reward functions but also to configure domain randomization parameters, enabling effective sim-to-real transfer.
ROSKA~\cite{ROSKA} incorporates both the best-performing policy and a randomly initialized policy in each iteration to maintain plasticity. Meanwhile, a Bayesian approach is used to guide policy selection.
R*~\cite{li2025r} further enhances Eureka’s performance through two mechanisms: structural evolution and parameter optimization. Structural evolution performs module-level crossover on high-quality reward functions to enable a more comprehensive search, while parameter optimization uses preference learning to fine-tune the parameters in the reward code. R* achieves consistently stronger results than Eureka.
However, these prior works largely overlook sample efficiency, which is a critical metric in RL. To address this, LaRes~\cite{li2025lares} integrates off-policy RL into the reward generation process by building a shared replay buffer, where each experience is labeled with rewards evaluated by the entire reward population.
In each generation, it inherits the best policy and value function, relabels the replay buffer, and applies reward scaling as well as constraint losses to avoid policy collapse. LaRes achieves superior sample efficiency and overall performance compared to Eureka, R*, and ROSKA.

Furthermore, several other works have investigated the design of reward functions under multi-agent settings.
LERO~\cite{wei2025lero} couples LLMs with an evolutionary outer loop to generate and refine hybrid reward functions (combining team and individual credit) and observation-enhancement functions; an inner MARL loop trains candidates and feeds back performance. On two tasks of MPE, LERO reports consistent gains and improved training efficiency.
ReMAC~\cite{li2025remac} targets multi-agent manipulation with a hierarchical design: an upper LLM-driven stage maintains and iteratively optimizes a population of team-level and agent-level rewards, while a lower stage applies MARL to learn collaborative policies. Experiments on ManiCraft benchmark and show that ReMAC-generated rewards outperform human-designed rewards.

\textbf{Morphological Evolution}: Morphological Evolution continuously optimizes the robot morphology and the control policy. In such problems, the final solution consists of two components: the optimal morphology and its associated policy. %This collaborative mode of EAs and RL differs from the mode defined in this section, where collaborative optimization refers to optimizing the same objective. 
In Morphological Evolution, EAs and RL optimize different aspects of the objective. We briefly introduce related work in this area. Classic algorithms in this category involve EAs for evolving morphologies and RL for learning policies~\cite{EvoGym}. Some works attempt to simultaneously optimize both morphology and policy using RL, such as CuCo~\cite{yuxin1} and Pre-Co~\cite{yuxin2}, or simultaneously employ EAs to optimize morphology and policy, such as HyperNEAT~\cite{DBLP:conf/ssci/TanakaA22}, or optimize morphology with other optimization techniques, such as Bayesian optimization~\cite{EvoGym}, {or explore the choice of genetic encoding for morphology~\cite{DBLP:conf/gecco/PigozziVM23}}. The hybrid algorithms in this area include EvoGym (GA)~\cite{EvoGym}, HERD~\cite{HERD}, AIEA~\cite{liu2023rapidly}, DERL~\cite{Fifei} and TAME~\cite{hejna2021task}. 

{
\textbf{Interpretable AI}: 
Policies derived from deep neural networks often lack interpretability, making them difficult to analyze and impractical for application in real-world scenarios with potential risks. 
To solve the problems, many works integrate decision trees with EAs and RL for highly interpretable policies.
Due to some methods lacking names in the original papers, we use abbreviations based on their characteristics to denote them.
POC-NLDT~\cite{dhebar2020interpretable} first collects a dataset using a policy pretrained with RL and then introduces two stages: open-loop training and closed-loop training.
During the open-loop training, optimization is performed using a bi-level EA~\cite{sinha2017review} based on the dataset. In this process, the upper level optimizes the tree structure, while the lower level seeks the optimal values for the tree's weights.
In the closed-loop training, further re-optimization of the weights is conducted using the cumulative reward collected across several episodes.
Finally, POC-NLDT demonstrates the interpretability and efficiency in four discrete action problems.
GE-QL~\cite{custode2022interpretable} evolves the tree structure using Grammatical Evolution (GE~\cite{DBLP:conf/eurogp/RyanCO98}) and optimizes leaf nodes using Q-learning.
CG-DT~\cite{custode2022interpretable} leverages GP to optimize structures of decision trees and employs CMA-ES~\cite{hansen1996adapting} to optimize weights.
CC-POC~\cite{custode2021co} extends POC-NLDT to continuous action spaces by constructing a population of actions. It utilizes GE for optimizing tree structures and UMDA$_{c}^G$~\cite{larranaga1999optimization} for action optimization. Q-learning is employed to optimize leaf nodes. CC-POC combines the two populations to obtain the complete solution.
QD-GT~\cite{ferigo2023quality} replaces GE with Map-Elites and defines behavioral descriptors based on action entropy and depth of decision trees. QD-GT demonstrates superior performance compared to GE schemes on the cart pole and mountain car tasks.
SIRL~\cite{lucio2024social} proposes a collaborative framework that constructs a population of decision trees. Actions are chosen randomly from the population to interact with the environment, and experiences are shared among them. Subsequently, each decision tree optimizes its leaf nodes using Q-learning and its tree structure using GE. SIRL demonstrates its efficiency across six MUJOCO tasks.
Besides, SVI~\cite{kubalik2021symbolic} is proposed to use Symbolic Regression to construct smooth analytical expression-based value functions, introducing symbolic value iteration to solve the Bellman equation. SVI offers higher interpretability compared to the black-box optimization of neural networks.
}

{
\textbf{Learning Classifier Systems}: 
Learning Classifier Systems (LCS)~\cite{urbanowicz2009learning} represent a class of methods that integrate learning with evolutionary principles to discover a set of rules capable of addressing a target task. LCS can also be referred to as population-based temporal-difference methods~\cite{XCSG}.
LCS consists of four key components: (1). A population of classifiers, representing the current knowledge base. Each classifier consists of a condition, an action, and an associated fitness parameter.
(2): A performance component, used to regulate the interaction between the environment and the population.
(3): A reinforcement component, allocates rewards obtained to classifiers.
(4): A discovery component, employed to discover new rules or refine existing ones.
LCS matches related classifiers based on inputs each time. If no match is found, random classifiers are generated and added to the population.
XCS~\cite{XCS} integrates Q-learning into LCS for learning, where each classifier represents an action-value function, and the associated parameters correspond to the weight matrix in function approximation. Each classifier includes a condition, an action, and four main parameters. XCS utilizes fitness based on accuracy, employing GA to search in the action space for classifier selection.
To improve the system's robustness and reduce parameter dependency, researchers introduce gradient descent methods into XCS, resulting in two approaches: XCS with direct gradient (XCSG) and XCS with residual gradient (XCSRG)~\cite{XCSG}.
XCSF evolves classifiers representing piecewise linear approximations of portions of the reward surface associated with the problem solution~\cite{lanzi2008learning}. In XCSF, the classifier’s prediction is calculated as a function, which is a linear combination of the classifier, rather than a scalar parameter. XCSF with tile coding~\cite{XCSFtile} replaces the original classifier prediction function in XCSF with a tile coding approximation. DGP-XCSF~\cite{DPGXCSF} employs graph-based dynamical genetic programming to represent traditional condition-action production system rules for solving continuous-valued reinforcement learning problems.
If you wish to explore further works related to LCS, you can refer to the work~\cite{urbanowicz2009learning}.}

\textbf{Challenges and Future Directions:}
% 初期的工作在协同优化中只能在部分任务中获得性能增益
Synergistic optimization has made significant progress in recent years. For instance, the early works based on the first collaboration approach are only able to achieve performance improvements on specific tasks~\cite{khadka2018evolution, khadka2019collaborative, bodnar2020proximal}. With the development of this field, recent works can consistently outperform both EAs and RL on a wide range of tasks~\cite{Re2}.
Despite the significant progress made in this direction, there remains a need for further investigation into how to effectively integrate the strengths of EAs and RL. 
The direction of synergistic optimization of EA and RL faces challenges similar to those of EA-assisted Optimization of RL and RL-assisted Optimization of EA, such as the need for domain knowledge, sensitivity to hyperparameters, and more. Concurrently, it presents a distinctive problem specific to this direction, namely, how to integrate EAs and RL to maximize the advantages of both for various problems.
Currently, this direction primarily focuses on SDP. 
Further research is required to explore how it can complement and provide advantages for addressing other types of problems.
In the future, researchers can explore in the following directions:
1) Explore how to integrate EAs and RL for synergistic optimization in addressing other problems.
2) Replace the foundational algorithms in current ERL methods with more advanced EAs or RL algorithms to fully leverage the advanced methods of both domains. 
3) For the first collaboration approach, design more efficient mechanisms where EAs influence RL or RL influences EA, enhancing the positive impacts between EAs and RL.
For the second collaboration approach, How to better and more automatically decompose problems to fully leverage their respective advantages is also an important research direction.
%4) Develop more efficient mechanisms for evaluating fitness to improve the sample efficiency of evolution.
In multi-agent settings, combining EAs and MARL is still at a nascent stage but holds substantial potential for advancement. 
Researchers can further delve into investigating how to fuse the capabilities of EAs and MARL to drive efficient collaboration.
%For example, cooperative evolution~\cite{ma2018survey} in EAs has the potential to address large-scale problems in multi-agent systems.
%In addition, combining EAs with MARL for competitive or cooperative-competitive hybrid problems is also a valuable research direction~\cite{wang2018exploiting, li2018punishment, wang2017onymity, wang2020communicating}.

%最初ERL相关工作只能在部分任务中展示出性能增益，随着ERL领域发展发展，最新的ERL工作在基本所有任务中都优于EA与RL。
%使用EA与RL用于并行优化已经显示出了强大的潜力，
%未来研究者们可以从以下几个方面进行进一步的研究。
% 一：使用更加前沿的EAs或者RL算法替代目前ERL算法中的基本算法，来充分利用两个领域最前沿的技术发展。
% 二：设计更加高效的EA影响RL，或者RL影响EA的机制，强化EA与RL正向影响。
% 三：涉及更好的适应度评估机制，来提升演化的样本效率。
% 而相较于单agent设定下的ERL，多agent下的ERL发展相对初期，有更多的发展空间与潜力，可以更多的从协作的角度来进行更好的EA与RL融合。

\section{Conclusion}

%  整体来说，通过整理与对比，我们发现EA辅助RL或者EA与RL协同优化的算法在时序优化问题中具有更显著的性能，其中EA与RL协同优化的工作带来的提升最为显著。而对于连续优化和组合优化问题，则更多的是RL辅助EA这类以EA为主的工作展现出更强大的能力。但同时这三个方向都有一个明显的不足，即没有一种结合方式被证明在不同类型的优化问题中都能有显著增益。

% Through careful analysis and comparison, we have found that algorithms involving EA-assisted RL or the synergistic optimization of EA and RL exhibit notably enhanced performance in sequential decision-making problems. In particular, synergistic optimization of EA and RL yields the most significant improvement, and the associated community is more active. Conversely, for continuous and combinatorial optimization problems, methods primarily centered on EA, such as RL-assisted EA, demonstrate more robust capabilities. However, all three directions share a conspicuous limitation: no integration approach has proven to yield significant gains across different types of optimization problems. 
% In addition, each research direction also faces different challenges. For example, in the case of EA-assisted RL, there are challenges related to additional overhead and the need for knowledge in the EA domain. RL-assisted EA introduces challenges related to hyperparameter sensitivity and the lack of systematic comparisons. These aspects require further research and resolution by researchers.

Overall, this paper systematically reviews different research directions within the field of ERL, along with the corresponding research branches within each direction. At the end of each section, the challenges faced by each direction and potential future research directions are summarized. We hope that this survey can comprehensively showcase the current development status of the ERL field to researchers, including existing algorithms, technical details, research challenges, and future research directions. 

It is worth emphasizing that, although this review indicates that EA-assisted RL and the synergistic optimization of EA and RL primarily focus on SDP, RL-assisted EA optimization is geared more towards addressing other optimization problems, closely tied to the problem-solving strengths of EAs and RL. However, these directions should not be limited to specific problems. Fortunately, in recent years, there has been a considerable amount of work attempting to leverage RL to solve problems where EAs excel, such as COP~\cite{DBLP:journals/cor/MazyavkinaSIB21}, or employing EAs to address SDP~\cite{kochenderfer2022algorithms}, yielding significant positive outcomes. This further broadens the application boundaries of different technologies. With the development of ERL, solutions to various problems will gradually emerge in different directions, which is a development we eagerly anticipate. Therefore, our survey primarily takes a technical perspective to assist researchers in thoughtful consideration and further expansion of the application boundaries in different research directions, driving the advancement of the field.

\nocite{}
\bibliographystyle{IEEEtran}
\bibliography{bare_jrnl_new_sample4}

\begin{IEEEbiography}
[{\includegraphics[width=1in,height=1.25in,clip,keepaspectratio]{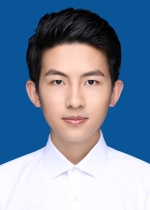}}] 
{Pengyi Li} received the B.S. degree from Tianjin University, China, in 2019. He is currently pursuing the Ph.D. degree in the College of Intelligence and Computing, Tianjin University. His current research interests are in reinforcement learning, evolutionary reinforcement learning, and multi-agent reinforcement learning.
\end{IEEEbiography}

\begin{IEEEbiography}
[{\includegraphics[width=1in,height=1.25in,clip,keepaspectratio]{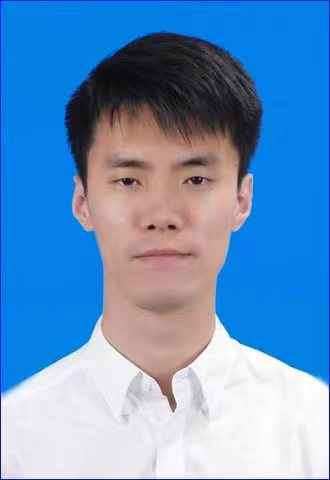}}] 
{Jianye Hao} (Senior Member, IEEE) is currently an
Associate Professor with Tianjin University,
Tianjin, China, and the Director of the Noah’s
Ark Decision-making and Reasoning Laboratory,
Huawei, Beijing, China. His research areas focus on
reinforcement learning and multiagent systems. 

The research of his team has been successfully applied
in domains such as game artificial intelligence
(AI), E-commerce recommendation, network
optimization, and supply chain optimization.
Prof. Hao has received a number of best paper
awards such as ASE2019 and CoRL2020.
\end{IEEEbiography}

\begin{IEEEbiography}
[{\includegraphics[width=1in,height=1.25in,clip,keepaspectratio]{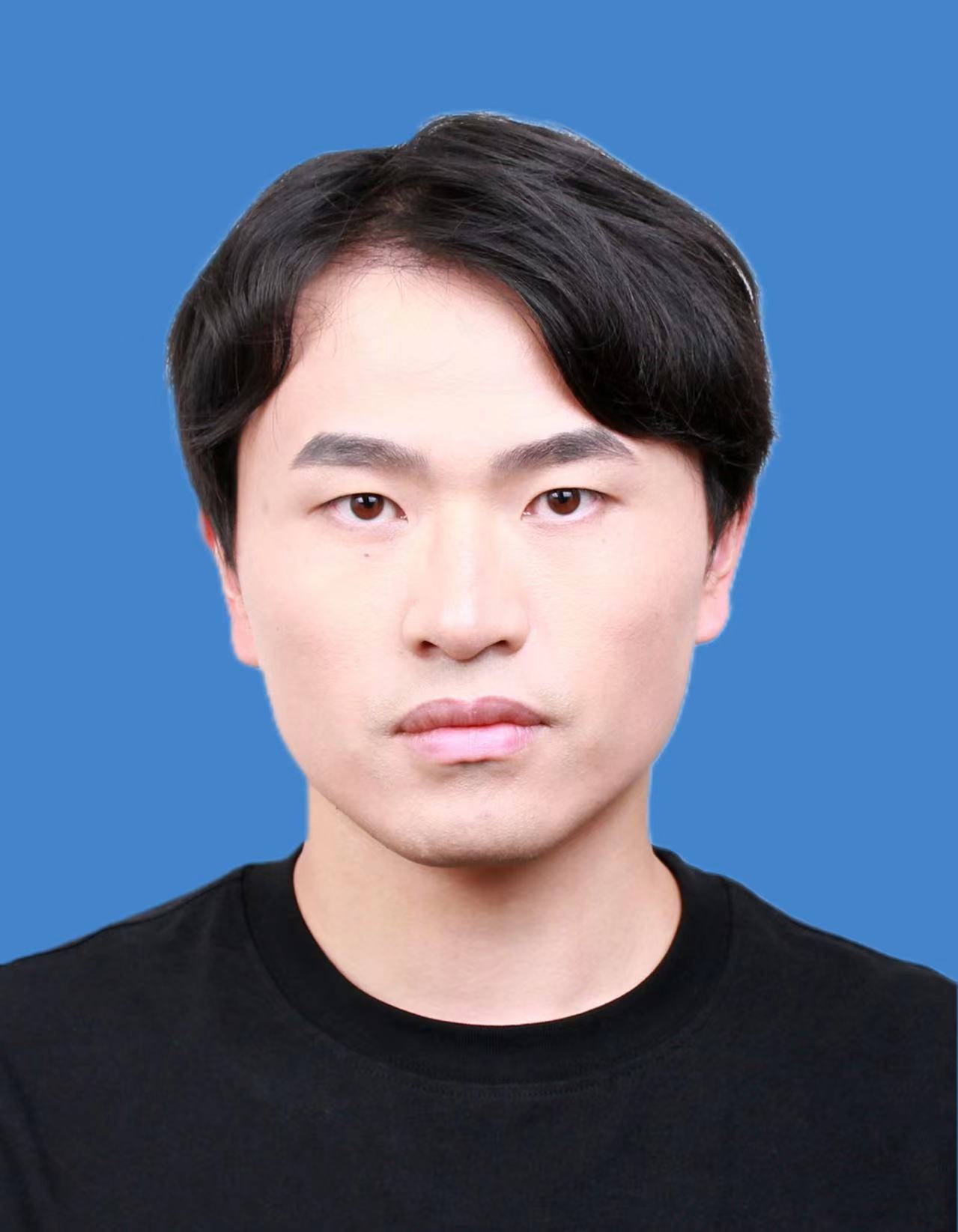}}] 
{Hongyao Tang} received the Ph.D. degree from
Tianjin University, Tianjin, China, in 2023. He currently 
holds a postdoctoral position at the Montreal Institute for Learning Algorithms (Mila) and the University of Montreal. His research interests include deep reinforcement learning, representation learning and continual learning.
\end{IEEEbiography}

\begin{IEEEbiography}
[{\includegraphics[width=1in,height=1.25in,clip,keepaspectratio]{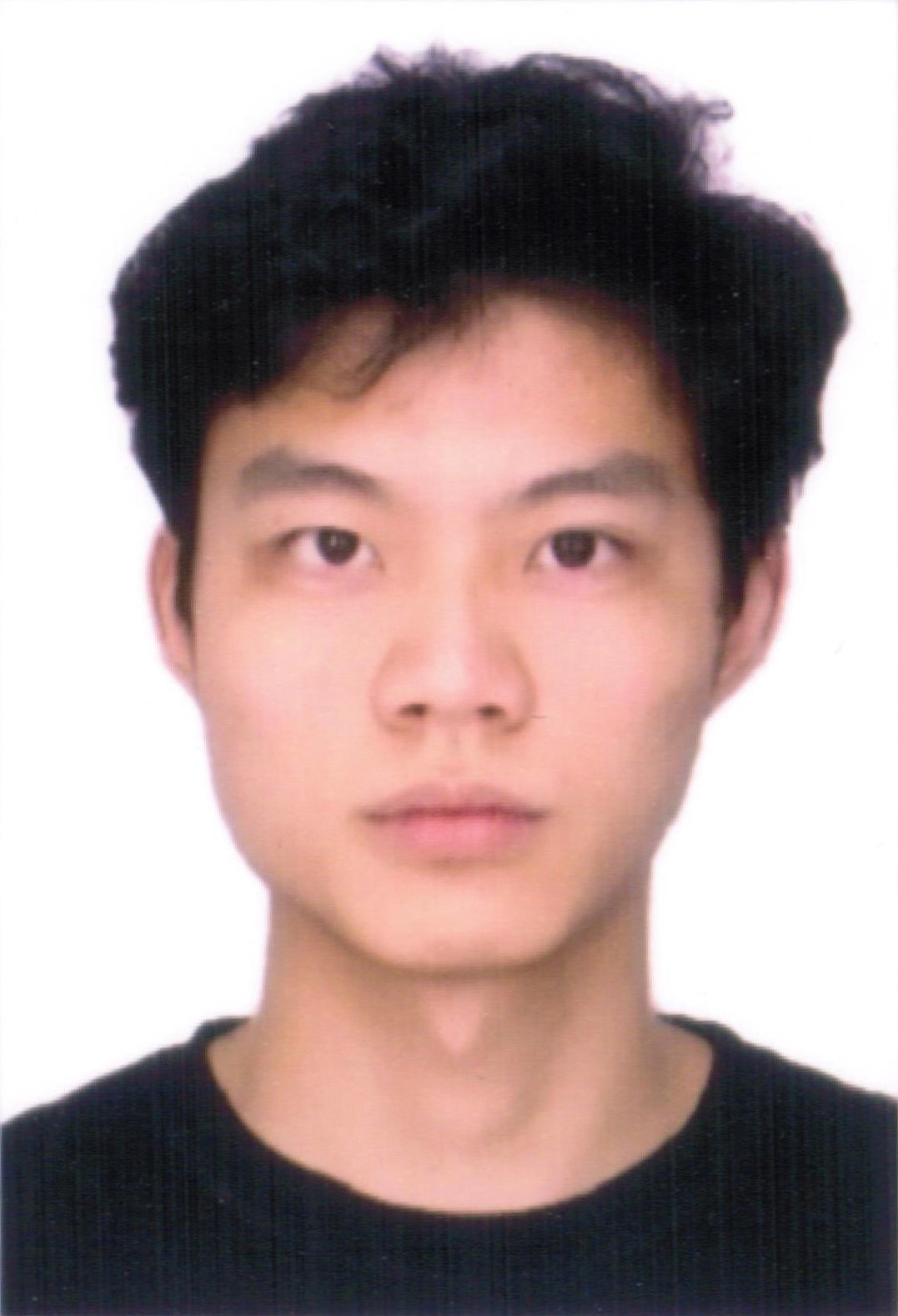}}] 
{Xian Fu} received the B.S. degree from Jilin University, Changchun, China, in 2022.
He is currently a master student with the College of Intelligence and Computing, Tianjin University. His current research interests are in evolutionary reinforcement learning, embodied AI, and multimodal language models.
\end{IEEEbiography}

\begin{IEEEbiography}
[{\includegraphics[width=1in,height=1.25in,clip,keepaspectratio]{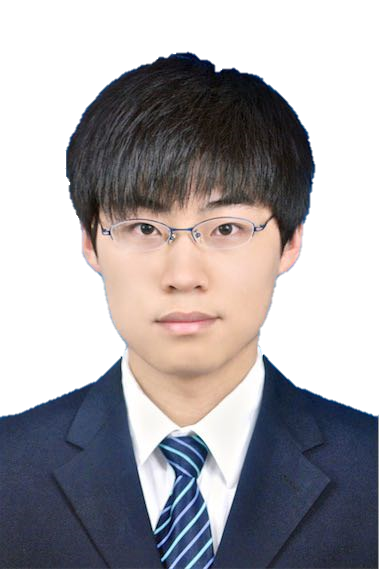}}] 
{Yan Zheng} is currently an Associate Professor with Tianjin University, Tianjin, China, focuses on the research and application of intelligent decision-making theories based on the integration of reinforcement learning and large models. In recent years, his work has successfully applied large language model-driven decision agents in multiple domains, including embodied control, bioinformatics, and data science.
\end{IEEEbiography}

\begin{IEEEbiography}
[{\includegraphics[width=1in,height=1.25in,clip,keepaspectratio]{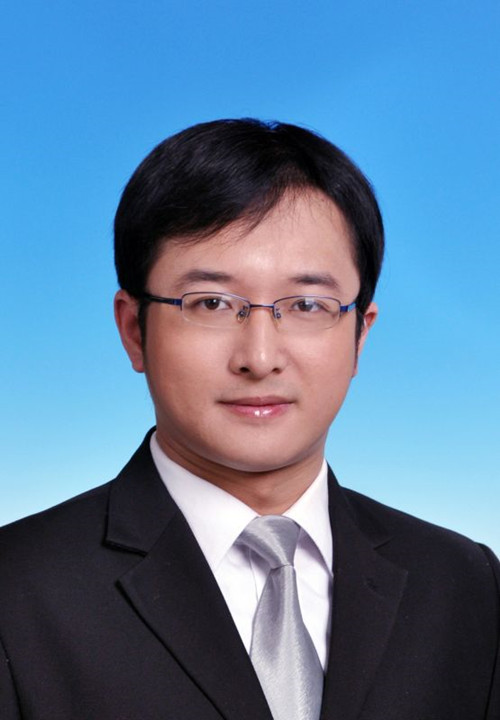}}] 
{Ke Tang} (Fellow, IEEE) received the the Ph.D.
degree from the Nanyang Technological University,
Singapore, in 2007. He is a Professor with the Department of Computer Science and Engineering, Southern University
of Science and Technology, Shenzhen, China. His
research interests mainly include evolutionary computation, machine learning, and their applications.

Dr. Tang was the recipient of the IEEE Computational Intelligence Society Outstanding Early Career Award and the
Natural Science Award of Ministry of Education (MOE) of China, and
was awarded the Newton Advanced Fellowship (Royal Society) and the
Chang jiang Professorship (MOE of China). He is an Associate Editor for the IEEE Transactions on Evolutionary Computation and a Member of Editorial Boards for a few other journals.
\end{IEEEbiography}

% \newpage

% \section{Biography Section}
% If you have an EPS/PDF photo (graphicx package needed), extra braces are
%  needed around the contents of the optional argument to biography to prevent
%  the LaTeX parser from getting confused when it sees the complicated
%  $\backslash${\tt{includegraphics}} command within an optional argument. (You can create
%  your own custom macro containing the $\backslash${\tt{includegraphics}} command to make things
%  simpler here.)
 
% \vspace{11pt}

% \bf{If you include a photo:}\vspace{-33pt}
% \begin{IEEEbiography}[{\includegraphics[width=1in,height=1.25in,clip,keepaspectratio]{fig1}}]{Michael Shell}
% Use $\backslash${\tt{begin\{IEEEbiography\}}} and then for the 1st argument use $\backslash${\tt{includegraphics}} to declare and link the author photo.
% Use the author name as the 3rd argument followed by the biography text.
% \end{IEEEbiography}

% \vspace{11pt}

% \bf{If you will not include a photo:}\vspace{-33pt}
% \begin{IEEEbiographynophoto}{John Doe}
% Use $\backslash${\tt{begin\{IEEEbiographynophoto\}}} and the author name as the argument followed by the biography text.
% \end{IEEEbiographynophoto}

\vfill

\end{document}